\documentclass[journal]{IEEEtran}
\PassOptionsToPackage{dvipsnames}{xcolor}
\usepackage[numbers,square,sort&compress]{natbib}

\usepackage[utf8]{inputenc} \usepackage[T1]{fontenc}    
\usepackage{hyperref}       \usepackage{url}            \usepackage{booktabs}       \usepackage{amsfonts}       \usepackage{nicefrac}       \usepackage{microtype}      
\usepackage{array}
\usepackage{ragged2e}

\newcolumntype{P}[1]{>{\raggedright\arraybackslash}p{#1}}

\usepackage[dvipsnames]{xcolor}
\usepackage[most]{tcolorbox}
\definecolor{SGDIE}{HTML}{3270CA}      \definecolor{ACCSGDIE}{HTML}{ED8838}   \colorlet{SGDIEbg}{SGDIE!15}
\colorlet{ACCSGDIEbg}{ACCSGDIE!18}

\DeclareRobustCommand{\sgdiehl}[1]{\ifmmode
    \tcboxmath[enhanced,colback=SGDIEbg,colframe=SGDIE,boxrule=0.4pt,left=2pt,right=2pt,top=0.4pt,bottom=0.4pt]{#1}\else
    \tcbox[on line,enhanced,fontupper=\small,colback=SGDIEbg,colframe=SGDIE,boxrule=0.4pt,left=2pt,right=2pt,top=0.4pt,bottom=0.4pt]{#1}\fi
}

\DeclareRobustCommand{\acchl}[1]{\ifmmode
    \tcboxmath[enhanced,colback=ACCSGDIEbg,colframe=ACCSGDIE,boxrule=0.4pt,left=2pt,right=2pt,top=0.4pt,bottom=0.4pt]{#1}\else
    \tcbox[on line,enhanced,fontupper=\small,colback=ACCSGDIEbg,colframe=ACCSGDIE,boxrule=0.4pt,left=2pt,right=2pt,top=0.4pt,bottom=0.4pt]{#1}\fi
}

\usepackage{placeins}
\usepackage{balance}
\usepackage{graphicx}
\usepackage[caption=false,font=normalsize,labelfont=sf,textfont=sf]{subfig}
\usepackage{textcomp}
\usepackage{stfloats}
\usepackage{makecell}
\usepackage{adjustbox}
\usepackage{multirow}
\usepackage{bm}
\usepackage{mathtools}
\usepackage{siunitx}
\usepackage{pifont}

\newcommand{\appref}[1]{\hyperref[#1]{Appendix~\ref*{#1}}}

\usepackage{amsmath}
\usepackage{amssymb}
\usepackage{amsthm}
\usepackage{aliascnt}

\usepackage[capitalize,noabbrev]{cleveref}

\theoremstyle{plain}

\newtheorem{assumption}{Assumption}
\newtheorem{theorem}{Theorem}[section]
\newaliascnt{proposition}{theorem}
\newtheorem{proposition}[proposition]{Proposition}
\aliascntresetthe{proposition}
\newaliascnt{lemma}{theorem}
\newtheorem{lemma}[lemma]{Lemma}
\aliascntresetthe{lemma}
\newaliascnt{corollary}{theorem}
\newtheorem{corollary}[corollary]{Corollary}
\aliascntresetthe{corollary}
\theoremstyle{definition}
\newaliascnt{remark}{theorem}

\aliascntresetthe{remark}

\definecolor{BrickRed}{RGB}{203, 65, 84}
\definecolor{OliveGreen}{RGB}{107, 142, 35}

\usepackage{colortbl}
\usepackage{enumitem}
\definecolor{mygray}{RGB}{226, 226, 226}
\definecolor{myred}{RGB}{252, 142, 142}
\definecolor{mygreen}{RGB}{147, 255, 143}
\definecolor{myblue}{RGB}{144, 155, 255}
\definecolor{myyellow}{RGB}{253, 253, 143}
\definecolor{mypurple}{RGB}{255, 142, 250}
\definecolor{softblue}{RGB}{100, 149, 237}
\definecolor{mygreen}{RGB}{0, 175, 175}
\definecolor{darkblue}{rgb}{0, 0, 0.5}
\definecolor{yxblue}{RGB}{0, 126, 189}
\hypersetup{
    pdftitle={Cross-Epoch Bias in SGD-based Leave-One-Out Influence: Analysis and Accumulative Correction},
    pdfauthor={Yunxiao Shi, Shuo Yang, Yixin Su, Rui Zhang, Min Xu},
    hidelinks
}

\begin{document}

\title{Cross-Epoch Bias in SGD-based Leave-One-Out Influence: Analysis and Accumulative Correction}
\author{Yunxiao~Shi, Shuo~Yang, Yixin~Su, Rui~Zhang, and Min~Xu%
\thanks{Corresponding author: Min Xu (e-mail: Min.Xu@uts.edu.au).}%
\thanks{Yunxiao Shi, Shuo Yang, and Min Xu are with the
University of Technology Sydney, Sydney, NSW, Australia.
E-mail: Yunxiao.Shi@student.uts.edu.au.}%
\thanks{Yixin Su is an independent researcher based in Wuhan, Hubei, China.
Website: \url{https://ethanmock.github.io/yixinsu.github.io}.}%
\thanks{Rui Zhang is an independent researcher based in Wuhan, Hubei, China.
Website: \url{https://www.ruizhang.info}.}}

\maketitle

\begin{abstract}
Training-data influence estimates support data selection and cleansing
by predicting the effects of removing individual examples.
The standard multi-epoch SGD-Influence Estimator (SGD-IE)
aggregates contributions from a sample's repeated appearances along
the recorded trajectory. When used to approximate finite deletion,
it omits an interaction: later removals act on the parameter
displacement accumulated from earlier removals.
We introduce the \textbf{Accumulative SGD-Influence Estimator
(ACC-SGD-IE)} to account for how earlier removals change the effect
of removing the same sample again. It uses the sample's Hessian
at subsequent appearances to correct the estimated parameter
change. Our non-convex analysis shows that this correction removes
SGD-IE's additional interaction contribution from the finite-horizon
error bound: the omitted interaction produces a local error linear
in the accumulated parameter change, while the remaining Taylor
remainder is bounded quadratically under Lipschitz-Hessian assumptions. The bounds characterize how training propagates the
remaining error. 
In non-convex experiments across clean and noisy tabular, text, and vision datasets, ACC-SGD-IE reduces validation-loss-change RMSE by $10.8\%$ and improves Jaccard@10\% by $15.2\%$ on average relative to SGD-IE. These gains also translate into lower test misclassification rates after data cleansing on MNIST and CIFAR10 at the evaluated removal budgets. On noisy industrial datasets, it reduces aggregate parameter RMS error with a mean relative reduction of $53.4\%$.
We further discuss computational tradeoffs, provide complementary
theoretical analysis and experimental evaluation for strongly convex
objectives, and explore potential extensions of the accumulative correction
mechanism to other influence estimators. \footnotemark[1]
\end{abstract}


\footnotetext[1]{Code is available at
\url{https://github.com/Ancientshi/ACC-SGD-IE}.}

\section{Introduction}\label{sec:Introduction}

\IEEEPARstart{M}{odern} deep learning depends on the quality and informativeness of its
training data. \emph{Leave-one-out (LOO) data influence} measures how
removing or perturbing one training example changes the learning
trajectory, its parameters, or its validation loss. It links model
behavior to individual training examples \citep{dattri}. These estimates
guide data-centric workflows: dataset cleansing and repair
\citep{hara2019data}, analysis of dataset structure and coverage
\citep{dataset_cartography}, and sample selection for efficient training
\citep{data_efficient_llmrec}. They support the broader shift toward
data-centric AI
\citep{yang2022dataset,beating_power_law,LESS,if_question,if_topn}.

Early work \citep{IF_regression,IF_ref1,pregibon1981logistic,distribution}
showed that, at the optimum of a convex loss, a sample's removal effect
can be inferred without retraining. \citet{koh2017understanding}
extended influence-function analysis \citep{influence_function} to
modern machine learning with smooth, strongly convex objectives.
Subsequent work found influence functions unreliable in the highly
non-convex objectives typical of deep networks
\citep{IF_Fragile,if_question}. \citet{hara2019data} addressed this issue
with SGD-IE, which follows the observed training trajectory to
approximate the parameter displacement caused by removing a sample.
Its multi-epoch form accumulates the contributions from all
occurrences of a sample, propagating each through subsequent
recorded updates.

\begin{figure}[!t]
    \centering
    \includegraphics[width=\linewidth]{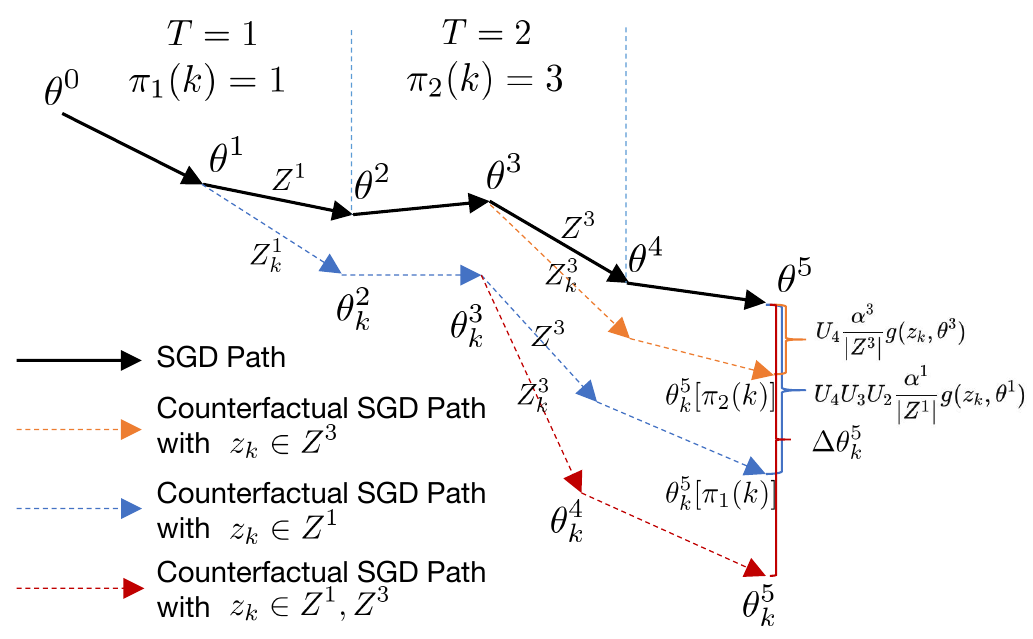}
    \caption{Illustrative finite-deletion mismatch across epochs in classical SGD-Influence Estimator. The red trajectory denotes the counterfactual with repeated deletion; the blue/orange paths illustrate independently injected linearized occurrence contributions. Their sum omits the interaction induced by a later deletion acting on the earlier displacement.}
    \label{fig:framework}
\end{figure}

In single-epoch training, removing a sample introduces one
perturbation. In multi-epoch training, earlier removals change
the parameters at which later removals occur. We show that,
when approximating finite deletion, \textbf{standard multi-epoch
SGD-IE omits the resulting adjustment to later removal
contributions.} We call the associated approximation discrepancy
\emph{cross-epoch accumulative bias} under the recorded training
schedule. \autoref{fig:framework} and \appref{app:bias}
illustrate this interaction using two deletion events.

To account for this interaction, we introduce the
\textbf{Accumulative SGD-Influence Estimator (ACC-SGD-IE)}.
Both SGD-IE and ACC-SGD-IE accumulate deletion contributions, but
they differ in how the accumulated displacement is propagated when
the target reappears. SGD-IE uses the full-batch curvature along the
recorded trajectory. ACC-SGD-IE adds an occurrence-level sample-Hessian
correction, propagating the displacement using the retained batch's
curvature. This incorporates the interaction between earlier and
later removals under the same recorded schedule and original batch
denominators.

Comparing their exact error recursions makes the effect of this
correction explicit. SGD-IE contains an additional deletion-interaction
residual that is linear in the true accumulated displacement.
ACC-SGD-IE incorporates this interaction into its propagation,
leaving the retained-gradient Taylor remainder, which is bounded
quadratically in the displacement under Lipschitz-Hessian assumptions.
Finite-horizon bounds characterize how non-convex training propagates
the remaining error. Our experiments then connect this analysis to
influence-estimation accuracy, ranking fidelity, and data cleansing,
while examining the computational cost of target-specific propagation.

We investigate three questions:
\emph{How does repeated deletion across epochs create a discrepancy
in the standard SGD-IE approximation?
How can this interaction be retained, and what determines the
remaining estimation error in non-convex training?
What gains does the correction provide over SGD-IE in influence
estimation, ranking, and data cleansing, and what computational
costs accompany them?}
We address these questions through estimator derivation, theoretical
error analysis, and empirical evaluation. Our contributions are
as follows:

\begin{itemize}[leftmargin=*]

\item \ding{182}
\textbf{\textit{Cross-epoch bias in SGD-based LOO influence.}}
We distinguish SGD-IE's infinitesimal sample-weight response from
finite repeated deletion under a fixed training schedule. We identify
the omitted interaction: a later removal acts on the parameter
displacement created by earlier removals. A two-occurrence example
makes this discrepancy explicit (\appref{app:bias}).

\item \ding{183}
\textbf{\textit{Accumulative SGD--Influence Estimator.}}
We derive ACC--SGD--IE from the coupled original and deletion updates.
Its sample-specific Hessian correction modifies propagation at target
re-occurrences, retaining the interaction between earlier and later
deletions while using derivatives along the recorded trajectory
(\autoref{sec:Accumulative SGD-Influence}).

\item \ding{184}
\textbf{\textit{Non-convex error analysis.}}
We derive exact error recursions that separate the deleted-sample
Hessian interaction from the retained-gradient Taylor remainder.
The correction removes the interaction residual, which is linear in
the true displacement; under Lipschitz-Hessian assumptions, the
remaining local remainder is bounded quadratically. Finite-horizon
bounds characterize how non-convex training propagates and may
amplify this error (\autoref{sec:error-analysis}).

\item \ding{185}
\textbf{\textit{Experiments and computational cost.}}
Across tabular, text, and vision benchmarks, experiments show gains
in loss-change estimation and influence ranking under clean and
corrupted training data, including improvements over extended
training trajectories. Data-cleansing experiments further demonstrate
downstream benefits at the evaluated removal budgets.
On industrial datasets, we compare estimated parameter and validation-loss
changes with those obtained by retraining. We also analyze and measure
the computation required for these estimates
(\autoref{subsec:exp_estimation},
\autoref{subsec:cross-epoch-fidelity},
\autoref{subsec:dataset_cleansing},
\autoref{subsec:industrial_noise}).

\end{itemize}

\section{Related Work}
\label{sec:Related Work}

\subsection{Data Influence and Estimation Method}

Data influence \citep{wang2025capturing,datainfluece_survey}, also called
data attribution \citep{dattri,SOURCE,TRAK}, measures how removing
\citep{koh2017understanding} or adding \citep{memorization_influence}
a training example changes the parameter trajectory or validation loss.
It provides per-example interpretability. For deep networks, repeatedly
retraining to compute these effects is infeasible. Influence estimators
therefore approximate them from a trained model or its trajectory.
Classical influence functions (IF) \citep{koh2017understanding} work
well in well-behaved convex regimes. SGD-IE \citep{hara2019data} instead
traces stochastic updates across epochs without requiring strong
convexity or convergence.

Later work focuses on scalability through checkpoints, projections,
and proxy models. Such approximations preserve coarse proportionality
useful for data-quality categorization \citep{dataset_cartography} and
scalable attribution \citep{TRAK,lpNTK}. Examples include TracIn
\citep{TracIn} and Adam-oriented methods such as LESS \citep{LESS}.
Recent LLM studies also examine how reliably IF-style methods transfer:
\citet{if_llm_work} report unstable and costly influence estimates on
LLMs, while \citet{chang2025scalable} identify a persistent mismatch
between causal influence and factual attribution in fact tracing.
These findings motivate work on both scalability and adaptation to
foundation models.

ACC-SGD-IE addresses the finite-deletion approximation within the
standard multi-epoch SGD-IE recursion. SGD-IE already propagates
information across epochs and computes the infinitesimal weight
derivative. Our correction accounts for how each later deletion acts
on the state changed by earlier deletions. The sample-specific
propagation operator retains this interaction, and the residual
analysis explains the resulting approximation error.

\subsection{Influence-based Data-Centric AI}

Data influence supports several data-centric AI applications
\citep{data_centric_AI,zha2023data}. In dataset cleansing,
influence estimates help identify mislabeled or noisy examples
whose removal can improve model performance
\citep{Data_cleaning,Raha,ridzuan2019review,hara2019data}.
Influence-guided selection also supports data-efficient LLM
fine-tuning
\citep{data_efficient_llmrec,LESS,DSDM,LLM_IF,jain2024data}.
More broadly, pruning and filtering large corpora can reduce
training cost and improve generalization
\citep{proxy,Data_Distribution_Search,xia2022moderate,coverage,yang2022dataset,beating_power_law,data_diet,forgetting_score,jain2024data,herding,coreset_survey,han-etal-2023-understanding,graph_data_selection}.
Other applications include active learning
\citep{active_ntk,active_learning_survey,select_if,Kding2018ActiveLF,donmez2008optimizing,roy2001toward},
unlearning \citep{CF_IF_unlearning,rec_unlearning_IF},
data poisoning and robustness analysis
\citep{influence_data_poisoning,perturbed_recsys,if_topn},
continual learning \citep{IF_CL}, and dataset distillation
\citep{datasetdistillation}.

These applications have different needs. Selecting or removing samples
from a large dataset requires fast estimates for many samples.
We study smaller models where more accurate estimates of individual
samples' effects may justify extra computation. Our industrial
experiments use defect-detection, equipment-monitoring, and sensor
datasets with label noise. We compare estimation accuracy and report
the computational cost.

\section{Preliminaries}
\label{sec:Preliminaries}

We consider a standard supervised deep-learning setting.
Let
\(
  D = \{z_j=(x_j,y_j)\}_{j=1}^n \subset \mathcal{X}\times\mathcal{Y}
\)
denote the training set, where \(x_j\in\mathcal{X}\) are inputs and
\(y_j\in\mathcal{Y}\) are associated targets (labels).
Examples are distinguished by their indices, even when their inputs and
labels coincide; removing \(z_k\) removes only record \(k\).
A deep neural network
\(
  f(\cdot;\theta):\mathcal{X}\to\mathcal{F},
  \quad \theta\in\Theta\subseteq\mathbb{R}^p,
\)
maps inputs to predictions in some output space \(\mathcal{F}\).
Given a non-negative loss
\(\ell:\mathcal{F}\times\mathcal{Y}\to\mathbb{R}_{\ge0}\),
we write the per-sample loss as
\(
  L(z,\theta)=\ell(f(x;\theta),y)
\)
and the empirical risk as
\(
  L(D,\theta)=\frac1n\sum_{j=1}^n L(z_j,\theta).
\)
We denote the per-sample gradient and Hessian by
\(
  g(z,\theta)=\nabla_\theta L(z,\theta)
\)
and
\(
  H(z,\theta)=\nabla_\theta^2 L(z,\theta),
\)
and the mini-batch Hessian by
\(
  H(Z,\theta)=\tfrac1{|Z|}\sum_{z\in Z}H(z,\theta).
\)

\subsection{SGD and SGD-Influence}
\paragraph{Stochastic Gradient Descent (SGD).}
For a run of $N$ updates, let \(\{Z^i\}_{i=0}^{N-1}\) be the mini-batch
sequence and \(\{\alpha^i\}_{i=0}^{N-1}\) the learning-rate schedule.
Each mini-batch contains a given record at most once. We use the same
recorded batch sequence and any other random choices for the ordinary
and deletion runs. This isolates the effect of removing the target
sample. The state \(\theta^i\) is the parameter
vector after \(i\) updates, for $i=0,\ldots,N$; \(Z^i\) is used for the
update from \(i\) to \(i+1\), for $i=0,\ldots,N-1$.
Starting from \(\theta^{0}\), mini-batch SGD performs
\begin{equation}
\label{eq:SGD}
  \theta^{i+1}
  \;\leftarrow\;
  \theta^{i}
  \;-\;
  \frac{\alpha^{i}}{|Z^{i}|}
  \sum_{z\in Z^{i}} g\bigl(z,\theta^{i}\bigr).
\end{equation}

\paragraph{Counterfactual SGD and SGD-Influence.}
Fix a training example \(z_k\in D\).
Our \emph{counterfactual SGD} masks record \(k\) in every original mini-batch,
starting from the same initialization \(\theta_k^0=\theta^0\) and keeping the
batch schedule, learning rates and original denominator \(|Z^i|\) fixed:
\begin{equation}
\label{eq:Counterfactual_SGD}
  \theta_k^{i+1}
  \leftarrow
  \theta_k^{i}
  -
  \frac{\alpha^{i}}{|Z^{i}|}
  \sum_{z\in Z_k^{i}} g\bigl(z,\theta_k^{i}\bigr),
  \qquad
  Z_k^{i}=Z^{i}\setminus\{z_k\}.
\end{equation}
The \emph{SGD-Influence} of \(z_k\) at step \(i\) is the parameter displacement
\begin{equation}
\label{eq:model_change}
  \Delta\theta_k^{i}
  \;:=\;
  \theta_k^{i}-\theta^{i},
\end{equation}
which measures the exact leave-one-out (LOO) parameter change under
this deletion procedure. We keep the original batch denominator so that
removing the target leaves every other sample's weight unchanged. For
example, deleting one sample from a batch of size $M$ leaves the retained
gradients divided by $M$. An empty retained batch produces no update.
Computing \(\Delta\theta_k^i\) by explicit retraining requires
a separate training run for each target, motivating efficient estimators.

\subsection{SGD-Influence Estimator}
\label{subsec:SGD-IE-Definition}
The SGD-Influence Estimator (SGD-IE) \citep{hara2019data} linearizes SGD
updates around the observed trajectory. It estimates how the parameters
respond to an infinitesimal reduction in a sample's weight. This
sample-weight tangent is also used to approximate the finite displacement
\(\Delta\theta_k^i\) caused by removing that sample. We recall its one-epoch
and multi-epoch forms before making this distinction precise.

\paragraph{One-epoch estimation.}
Let \(z_k\) appear once in an epoch, in mini-batch \(Z^{\pi(k)}\).
Define the step-wise transition matrix
\(
  U_i := I - \alpha^i H(Z^i,\theta^i),\quad i=0,\ldots,N-1.
\)
Under a single-epoch schedule—each example is seen exactly once—the influence
of removing \(z_k\) on the parameters after \(i\) updates is approximated by
\begin{equation}
\label{eq:One_Epoch_SGD_Influence_Estimation}
\begin{aligned}
  \widehat{\Delta\theta_k^{i}}
  &=
  \frac{\alpha^{\pi(k)}}{|Z^{\pi(k)}|}
  \,U_{i-1}U_{i-2}\cdots U_{\pi(k)+1}\,
  \\
  &\qquad {}\cdot g\bigl(z_k,\theta^{\pi(k)}\bigr),
  \qquad i>\pi(k),
\end{aligned}
\end{equation}
which propagates the instantaneous perturbation at step \(\pi(k)\) through all
subsequent updates in that epoch.
The product is the identity when \(i=\pi(k)+1\).

\paragraph{Multi-epoch estimation.}
Now let training run for \(T\) complete epochs with \(B\) mini-batches each,
with every record appearing exactly once per epoch, so the total number
of steps is \(N=TB\). Write
\(0\le\pi_1(k)<\dots<\pi_T(k)<N\) for the indices at which \(z_k\) appears.
The canonical multi-epoch SGD-Influence Estimator aggregates one-epoch
surrogates across all occurrences:
\begin{equation}
\label{eq:SGD_Influence_Estimation}
\begin{aligned}
  \widehat{\Delta\theta_k^{i}}
  &=
  \sum_{t:\,\pi_t(k)<i}
    \frac{\alpha^{\pi_t(k)}}{|Z^{\pi_t(k)}|}
    \,U_{i-1}U_{i-2}\cdots U_{\pi_t(k)+1}\,
    \\
  &\qquad {}\cdot g\bigl(z_k,\theta^{\pi_t(k)}\bigr).
\end{aligned}
\end{equation}

To distinguish a small weight change from full removal, let
$\varepsilon$ be the fraction of the target's contribution suppressed
at every occurrence. A value of $0.01$ reduces its weight by $1\%$;
$\varepsilon=1$ removes it completely. The next result shows that
SGD-IE computes the slope at $\varepsilon=0$.

\begin{proposition}[SGD-IE as a sample-weight tangent]
\label{prop:sgdie-weight-tangent}
Let \(\theta_k^i(\varepsilon)\) follow the same initialization and coupled
updates with the contribution of record \(k\) multiplied by
\(1-\varepsilon\), retaining the original batch denominator. For twice
differentiable per-sample losses and differentiable update paths, we have
\[
  \left.\frac{d\theta_k^i(\varepsilon)}{d\varepsilon}\right|_{\varepsilon=0}
  =\widehat{\Delta\theta_k^i}.
\]
The finite displacement is instead
\(\Delta\theta_k^i=\theta_k^i(1)-\theta_k^i(0)\).
\end{proposition}
Here \(\varepsilon=0\) gives ordinary training and \(\varepsilon=1\) gives
full removal. SGD-IE is the local slope at the ordinary trajectory;
finite deletion measures the change across this entire interval. The
proof appears in \appref{app:weight-tangent-proof}.
During multi-epoch training, an earlier removal changes the parameters
at which the target's later appearances are evaluated. These
repeated-deletion interactions are absent from the small-weight-change aggregation
in \autoref{eq:SGD_Influence_Estimation}.

\section{Accumulative SGD-Influence Estimator}
\label{sec:Accumulative SGD-Influence}

ACC-SGD-IE tracks how the effect of deleting a sample evolves throughout
training. The first removal changes the parameters. When the same sample
appears again, its removal acts on parameters already changed by the
earlier removal. ACC-SGD-IE accounts for this interaction through a
sample-specific Hessian correction (\autoref{eq:define_V}).

Let $\Delta\theta_k^i=\theta_k^i-\theta^i$ be the exact fixed-denominator
leave-one-out displacement after $i$ SGD updates, with
\[
  U_i \;=\; I \;-\;\alpha^i\,H\bigl(Z^i,\theta^i\bigr),
  \qquad
  \Delta\theta_k^{0}=0.
\]
At each update, the parameter change has two contributions: the effect
carried forward from earlier updates and the new effect of removing the
target from the current batch. The matrix $V_i^k$ propagates the former,
and the vector $\Gamma_{k,i}$ supplies the latter. Subtracting the
ordinary and counterfactual updates gives the exact recurrence
\begin{equation}\label{eq:recursive_formula}
  \Delta\theta_k^{i+1} \;=\;
  V_i^k\,\Delta\theta_k^{\,i}
  \;+\;
  \Gamma_{k,i}
  \;-\;\alpha^i\widetilde R_k^i,
\end{equation}
where $V_i^k$ and $\Gamma_{k,i}$ are defined for $i=0,\ldots,N-1$ by
\begin{align}
  V_i^k 
  &= U_i 
    + {\textcolor{purple}{\bm{\mathbb{I}(z_k\in Z^i)\,
      \frac{\alpha^i}{|Z^i|}\,
      H\bigl(z_k,\theta^i\bigr)}}}
  \,,\label{eq:define_V}\\
\Gamma_{k,i}
  &= \mathbb{I}(z_k\in Z^i)\,
      \frac{\alpha^{\,i}}{|Z^{i}|}\,
      g\bigl(z_k,\theta^{i}\bigr)
  \,,\label{eq:define_Gamma}
\end{align}
The full-batch transition $U_i$ includes the curvature of every sample,
including the target. The correction in $V_i^k$ adds back the target
term that $U_i$ had subtracted, leaving only the retained samples'
curvature in the update. Earlier parameter changes are therefore
propagated using the retained batch. The vector
$\Gamma_{k,i}$ is the gradient contribution omitted when the target is
removed. Both corrections are active only when $z_k\in Z^i$.

To obtain this recurrence, we expand the retained gradients around the
original parameters $\theta^i$. The Taylor residual collects the part
of their change beyond this linear approximation:
\[
  \widetilde R_k^i
  :=\frac1{|Z^i|}\sum_{z\in Z^i\setminus\{z_k\}}
  \bigl[g(z,\theta_k^i)-g(z,\theta^i)
        -H(z,\theta^i)\Delta\theta_k^i\bigr].
\]
If the retained-sample Hessians change by at most $L$ times the parameter
distance along the segments connecting the two trajectories, the
residual norm is at most
$\frac L2\|\Delta\theta_k^i\|_2^2$.
ACC-SGD-IE uses the linear approximation, giving the estimator update
\[
  \widetilde{\Delta\theta}_k^{i+1}
  =V_i^k\widetilde{\Delta\theta}_k^i+\Gamma_{k,i},
  \qquad \widetilde{\Delta\theta}_k^0=0.
\]
Let $\pi_t(k)$ index the $t$-th occurrence of $z_k$. Unrolling this
estimator recurrence yields our \textbf{Accumulative SGD-Influence Estimator
(ACC-SGD-IE)}:
\begin{equation}\label{eq:estimator}
  \boxed{
\widetilde{\Delta\theta}_k^i
\;:=\;
\sum_{t:\,\pi_t(k)<i}
  \Bigl(
    V_{i-1}^k V_{i-2}^k\cdots V_{\pi_t(k)+1}^k
  \Bigr)
  \Gamma_{k,\pi_t(k)}
    }.
\end{equation}
Each summand starts with the gradient contribution from one occurrence
of the target. The matrix product carries that contribution through all
later updates, including subsequent removals of the same target.
Thus ACC-SGD-IE retains the interaction between removals while using a
linear approximation of each retained gradient.

\section{Error Analysis and Comparison}
\label{sec:error-analysis}

We analyze where estimation errors arise and how later updates propagate
them. The analysis applies to smooth non-convex objectives, including
neural-network losses. 

Let $M=|Z^i|$ be the fixed batch size. The indicator
$\iota_i^k=\mathbb{I}(z_k\in Z^i)$ records whether update $i$ contains
the target, and $c_i^k=1-\iota_i^k/M$ is the fraction of the batch retained
after its removal. Write
$U_i=I-\alpha^i H(Z^i,\theta^i)$ and
$V_i^k=U_i+\iota_i^k\alpha^i H(z_k,\theta^i)/M$.
These matrices propagate parameter changes through the full-batch and
retained-batch updates, respectively. The true parameter change caused
by deletion is $\Delta\theta_k^i=\theta_k^i-\theta^i$, with magnitude
$d_i^k=\|\Delta\theta_k^i\|_2$. The estimation errors are
$E_k^i=\widehat{\Delta\theta_k^i}-\Delta\theta_k^i$ and
$\widetilde E_k^i=\widetilde{\Delta\theta}_k^i-\Delta\theta_k^i$
for SGD-IE and ACC-SGD-IE, respectively.

\subsection{Sources and propagation of estimation error}
Each update carries forward the previous estimation error and adds an
error from approximating the gradient change. We call the gradient
mismatch a local residual; multiplying it by the learning rate gives
the new parameter error. Unrolling the recurrence sums these
contributions after accounting for the later updates that act on each one.
For SGD-IE, the exact recursion is
\begin{equation}\label{eq:error-recursion-classic-main}
\begin{aligned}
 E_k^{i+1}&=U_iE_k^i+\alpha^i R_k^i,
 \\
 E_k^N&=\sum_{i=0}^{N-1}\alpha^i\Phi_{N-1,i+1}R_k^i,\\
 R_k^i&:=\frac1M\sum_{z\in Z^i\setminus\{z_k\}}
 [g(z,\theta_k^i)-g(z,\theta^i)]\\
 &\qquad {}-H(Z^i,\theta^i)\Delta\theta_k^i.
\end{aligned}
\end{equation}
Here $\Phi_{j,i}=U_jU_{j-1}\cdots U_i$ represents propagation from
updates $i$ through $j$. The latest matrix is on the left, and an empty
product is the identity.
For ACC-SGD-IE, the corresponding recursion is
\begin{equation}\label{eq:acc-error-rec-main}
\begin{aligned}
 \widetilde E_k^{i+1}&=V_i^k\widetilde E_k^i+\alpha^i\widetilde R_k^i,
 \\
 \widetilde E_k^N&=\sum_{i=0}^{N-1}
 \alpha^i\widetilde\Phi_{N-1,i+1}^k\widetilde R_k^i,\\
 \widetilde R_k^i&:=\frac1M\sum_{z\in Z^i\setminus\{z_k\}}
 \bigl[g(z,\theta_k^i)-g(z,\theta^i)\\
 &\qquad\qquad {}-H(z,\theta^i)\Delta\theta_k^i\bigr],
\end{aligned}
\end{equation}
where $\widetilde\Phi_{j,i}^k=V_j^kV_{j-1}^k\cdots V_i^k$.
The expression for $\widetilde R_k^i$ compares the true change in each
retained gradient with its linear approximation. The two local residuals
differ by one additional term:
\[
 R_k^i=\widetilde R_k^i-
 \frac{\iota_i^k}{M}H(z_k,\theta^i)\Delta\theta_k^i.
\]
This term captures the repeated-deletion interaction: when the target
reappears, its Hessian acts on the parameter change accumulated from
earlier removals. ACC-SGD-IE incorporates this interaction into $V_i^k$,
leaving only the retained-gradient Taylor remainder as its local
residual. If the retained per-sample Hessians are $L$-Lipschitz along
the segment joining $\theta^i$ and $\theta_k^i$, this remainder satisfies
$\|\widetilde R_k^i\|_2\le c_i^k L(d_i^k)^2/2$.
\appref{app:error-recursions} derives both recursions step by step.

\subsection{Error bounds for non-convex training}
The following theorem compares the estimation errors of SGD-IE and
ACC-SGD-IE in non-convex training. The bounds separate the
repeated-deletion interaction from the retained-gradient Taylor
remainder. Full derivations appear in
\appref{sec:noconvex-case}. We also analyze the smooth strongly convex objectives provided in \appref{sec:convex-case}.

\begin{theorem}[Non-convex error comparison]
\label{thm:nonconvex}
Suppose the relevant per-sample Hessians have norm at most
$\Lambda>0$ and are $L$-Lipschitz along the segments joining the
original and counterfactual iterates. Assume also that
$\|g(z_k,\theta^i)\|_2\le G$ at each occurrence of the target on
the original trajectory. Let $M\ge1$ and $\alpha^i\ge0$, and define
\[
 A_i=\sum_{j=0}^{i-1}\alpha^j,
 \qquad x=\Lambda A_N.
\]
Under \autoref{asm:assumption-sgd-nonconvex}, the displacement
satisfies
\[
 d_i^k\le
 \frac{G}{M\Lambda}\bigl(e^{\Lambda A_i}-1\bigr).
\]
The estimation errors satisfy
\[
\sgdiehl{
 \begin{aligned}
 \|E_k^N\|_2&\le
 \frac{G}{M^2\Lambda}\bigl(xe^x-e^x+1\bigr)
 \\
 &\quad {}+\frac{LG^2}{2M^2\Lambda^3}
 \bigl(e^{2x}-1-2xe^x\bigr),
 \end{aligned}
}
\]
and
\[
\acchl{
 \|\widetilde E_k^N\|_2\le
 \frac{LG^2}{2M^2\Lambda^3}
 \bigl(e^{2x}-1-2xe^x\bigr).
}
\]
More precise trajectory-dependent bounds, retaining the actual
target occurrences and products of propagation norms, are given
in \autoref{thm:nonconvex-path-bounds}.
\end{theorem}

\noindent\textbf{Interpretation.}
The SGD-IE upper bound contains two contributions: the
repeated-deletion interaction and the retained-gradient Taylor
remainder. ACC-SGD-IE incorporates the interaction into $V_i^k$,
removing the corresponding local residual and leaving only the
Taylor-remainder contribution in its upper bound. Thus, the
correction addresses the additional residual source in SGD-IE
under the stated non-convex assumptions.

\begin{table*}[!t]
  \centering
  \caption{Comprehensive comparison of influence–estimation accuracy and robustness under non-convex regime. The best value for each metric within a setting is \textbf{bold}. We add i.i.d.\ noise $x \!\leftarrow\! x+\mathcal{N}(0,\sigma^{2})$ to every feature on Adult \& MNIST. We randomly set a fraction $\sigma\times100\%$ of the zero entries in each one-hot word vector to 1 on 20News. A fraction $\rho\times 100\%$ of the training labels is flipped uniformly at random.}
  \label{tab:all_influence}
  \resizebox{\textwidth}{!}{\sisetup{detect-all}
  \begin{tabular}{@{}l c l l l l l l l@{}}
    \specialrule{1pt}{0pt}{0pt}
    \textbf{Dataset} & \textbf{Setting} & \textbf{Method} &
    {\textbf{RMSE}$\;\downarrow$} &
    {\textbf{Kendall's Tau}$\;\uparrow$} &
    {\textbf{Jacc@70\%}$\;\uparrow$} &
    {\textbf{Jacc@50\%}$\;\uparrow$} &
    {\textbf{Jacc@30\%}$\;\uparrow$} &
    {\textbf{Jacc@10\%}$\;\uparrow$} \\
    \midrule[0.8pt]
    \multicolumn{9}{c}{\textbf{Part 1.\; Estimation on Clean Data.}}\\
    \midrule[0.8pt]
    \multirow{2}{*}{Adult}
& \multirow{2}{*}{--} & SGD-IE & $2.40(1.10)\times10^{-3}$ & 0.4064(0.13) & 0.6059(0.05) & 0.4868(0.09) & 0.4331(0.11) & 0.3629(0.09) \\
& & ACC-SGD-IE & \multicolumn{1}{l}{$\bm {2.20(1.10)\times10^{-3}}$} & \textbf{0.4220(0.13)} & \textbf{0.6080(0.05)} & \textbf{0.4914(0.09)} & \textbf{0.4388(0.11)} & \textbf{0.3834(0.09)} \\

    \midrule
    \multirow{2}{*}{20News}
& \multirow{2}{*}{--} & SGD-IE & $1.82(1.41)\times10^{-2}$ & 0.2248(0.16) & 0.5582(0.02) & 0.3696(0.04) & 0.2424(0.07) & 0.1403(0.08) \\
& & ACC-SGD-IE & \multicolumn{1}{l}{$\bm {1.78(1.48)\times10^{-2}}$} & \textbf{0.2632(0.16)} & \textbf{0.5682(0.02)} & \textbf{0.3884(0.06)} & \textbf{0.2489(0.08)} & \textbf{0.1589(0.10)} \\

    \midrule
    \multirow{2}{*}{MNIST}
& \multirow{2}{*}{--} & SGD-IE & $3.40(6.10)\times10^{-3}$ & 0.4009(0.15) & \textbf{0.6095(0.04)} & 0.4794(0.07) & 0.4019(0.12) & 0.4129(0.15) \\
& & ACC-SGD-IE & \multicolumn{1}{l}{$\bm {2.00(2.70)\times10^{-3}}$} & \textbf{0.4058(0.15)} & 0.6087(0.04) & \textbf{0.4796(0.07)} & \textbf{0.4057(0.12)} & \textbf{0.4297(0.16)} \\
    \midrule
    \multicolumn{3}{c}{\textbf{Average Improve (\%)}} 
& \multicolumn{1}{c}{$\downarrow17.24\%$}
& \multicolumn{1}{c}{$\uparrow7.38\%$}
& \multicolumn{1}{c}{$\uparrow0.67\%$}
& \multicolumn{1}{c}{$\uparrow2.02\%$}
& \multicolumn{1}{c}{$\uparrow1.65\%$}
& \multicolumn{1}{c}{$\uparrow7.66\%$} \\
    \midrule[0.8pt]
    \multicolumn{9}{c}{\textbf{Part 2.\; Estimation on Data Corrupted by Feature Noise.}}\\
    \midrule[0.8pt]
    \multirow{4}{*}{Adult}
        & \multirow{2}{*}{$\sigma=0.01$}

& SGD-IE & $3.80(3.20)\times10^{-3}$ & 0.3016(0.14) & \textbf{0.5861(0.03)} & 0.4465(0.05) & 0.3679(0.08) & 0.3092(0.09) \\
& & ACC-SGD-IE & \multicolumn{1}{l}{$\bm {3.70(3.20)\times10^{-3}}$} & \textbf{0.3352(0.14)} & 0.5818(0.03) & \textbf{0.4594(0.06)} & \textbf{0.3844(0.09)} & \textbf{0.3354(0.10)} \\

        \cmidrule{2-9}
        & \multirow{2}{*}{$\sigma=0.05$}
& SGD-IE & $3.00(1.10)\times10^{-3}$ & 0.2636(0.09) & 0.5775(0.02) & \textbf{0.4386(0.03)} & 0.3410(0.05) & 0.2335(0.09) \\
& & ACC-SGD-IE & \multicolumn{1}{l}{$\bm {2.80(1.00)\times10^{-3}}$} & \textbf{0.2765(0.10)} & \textbf{0.5800(0.02)} & 0.4335(0.03) & \textbf{0.3556(0.06)} & \textbf{0.2927(0.08)} \\
    \midrule
    \multirow{2}{*}{20News}
        & \multirow{2}{*}{$\sigma=2.5\times10^{-3}$}
& SGD-IE & $2.02(0.81)\times10^{-2}$ & 0.0521(0.05) & 0.5427(0.04) & 0.3661(0.05) & 0.2505(0.02) & 0.1270(0.01) \\
& & ACC-SGD-IE & \multicolumn{1}{l}{$\bm {1.74(0.69)\times10^{-2}}$} & \textbf{0.1114(0.08)} & \textbf{0.5799(0.00)} & \textbf{0.3847(0.01)} & \textbf{0.2866(0.00)} & \textbf{0.1438(0.03)} \\
    \midrule
    \multirow{4}{*}{MNIST}
        & \multirow{2}{*}{$\sigma=0.01$}
& SGD-IE & $9.00(5.00)\times10^{-4}$ & 0.1659(0.11) & 0.5661(0.02) & 0.3912(0.03) & 0.2845(0.05) & 0.2656(0.07) \\
& & ACC-SGD-IE & \multicolumn{1}{l}{$\bm {8.00(4.00)\times10^{-4}}$} & \textbf{0.2023(0.11)} & \textbf{0.5689(0.02)} & \textbf{0.4059(0.05)} & \textbf{0.3131(0.06)} & \textbf{0.3067(0.09)} \\
        \cmidrule{2-9}
        & \multirow{2}{*}{$\sigma=0.05$} 
& SGD-IE & $2.70(1.30)\times10^{-3}$ & 0.1348(0.10) & 0.5535(0.02) & 0.3712(0.03) & 0.2512(0.04) & 0.2276(0.08) \\
& & ACC-SGD-IE & \multicolumn{1}{l}{$\bm {1.30(0.90)\times10^{-3}}$} & \textbf{0.1894(0.09)} & \textbf{0.5565(0.02)} & \textbf{0.3868(0.03)} & \textbf{0.2848(0.05)} & \textbf{0.3026(0.09)} \\
    \midrule
    \multicolumn{3}{c}{\textbf{Average Improve (\%)}} 
    & \multicolumn{1}{c}{$\downarrow17.22\%$} 
    & \multicolumn{1}{c}{$\uparrow38.46\%$} 
    & \multicolumn{1}{c}{$\uparrow1.52\%$} 
    & \multicolumn{1}{c}{$\uparrow2.95\%$} 
    & \multicolumn{1}{c}{$\uparrow9.32\%$} 
    & \multicolumn{1}{c}{$\uparrow19.10\%$} \\
    \midrule[0.8pt]
    \multicolumn{9}{c}{\textbf{Part 3.\; Estimation on Data Corrupted by Label Noise.}}\\
    \midrule[0.8pt]
    \multirow{4}{*}{Adult}
        & \multirow{2}{*}{$\rho = 0.1$}
& SGD-IE & $2.00(1.30)\times10^{-3}$ & 0.5003(0.14) & 0.6258(0.06) & 0.5084(0.09) & 0.4606(0.12) & 0.4061(0.10) \\
& & ACC-SGD-IE & \multicolumn{1}{l}{$\bm {2.00(1.30)\times10^{-3}}$} & \textbf{0.5035(0.13)} & \textbf{0.6259(0.06)} & \textbf{0.5164(0.09)} & \textbf{0.4701(0.11)} & \textbf{0.4432(0.09)} \\

        \cmidrule{2-9}
        & \multirow{2}{*}{$\rho = 0.3$}
& SGD-IE & $1.50(0.80)\times10^{-3}$ & 0.6330(0.12) & 0.6793(0.08) & 0.5833(0.10) & 0.5130(0.13) & \textbf{0.4270(0.13)} \\
& & ACC-SGD-IE & \multicolumn{1}{l}{$\bm {1.50(0.80)\times10^{-3}}$} & \textbf{0.6362(0.12)} & \textbf{0.6794(0.08)} & \textbf{0.5849(0.10)} & \textbf{0.5185(0.13)} & 0.4256(0.13) \\
    \midrule
    \multirow{4}{*}{20News}
        & \multirow{2}{*}{$\rho = 0.1$}
& SGD-IE & $1.13(2.23)\times10^{-2}$ & 0.2728(0.12) & 0.5587(0.02) & 0.3784(0.04) & 0.2648(0.03) & 0.1600(0.05) \\
& & ACC-SGD-IE & \multicolumn{1}{l}{$\bm {1.08(1.48)\times10^{-2}}$} & \textbf{0.3123(0.10)} & \textbf{0.5629(0.02)} & \textbf{0.3973(0.04)} & \textbf{0.2895(0.04)} & \textbf{0.2124(0.06)} \\
        \cmidrule{2-9}
        & \multirow{2}{*}{$\rho = 0.3$} 
& SGD-IE & $1.35(0.83)\times10^{-2}$ & 0.3004(0.14) & 0.5694(0.03) & 0.4010(0.05) & 0.2670(0.04) & 0.1792(0.07) \\
& & ACC-SGD-IE & \multicolumn{1}{l}{$\bm {1.30(0.80)\times10^{-2}}$} & \textbf{0.3329(0.14)} & \textbf{0.5816(0.04)} & \textbf{0.4164(0.06)} & \textbf{0.2890(0.06)} & \textbf{0.2293(0.08)} \\

    \midrule
    \multirow{4}{*}{MNIST}
        & \multirow{2}{*}{$\rho = 0.1$}
& SGD-IE & $1.80(3.61)\times10^{-2}$ & 0.1832(0.07) & \textbf{0.5528(0.02)} & 0.3739(0.03) & 0.2581(0.04) & 0.2206(0.07) \\
& & ACC-SGD-IE & \multicolumn{1}{l}{$\bm {1.74(3.72)\times10^{-2}}$} & \textbf{0.1935(0.08)} & 0.5521(0.02) & \textbf{0.3750(0.03)} & \textbf{0.2696(0.04)} & \textbf{0.2430(0.08)} \\
        \cmidrule{2-9}
        & \multirow{2}{*}{$\rho = 0.3$}
& SGD-IE & $4.95(3.04)\times10^{-2}$ & 0.1384(0.08) & 0.5444(0.01) & 0.3535(0.02) & 0.2059(0.03) & 0.1064(0.04) \\
& & ACC-SGD-IE & \multicolumn{1}{l}{$\bm {4.88(3.01)\times10^{-2}}$} & \textbf{0.1491(0.07)} & \textbf{0.5468(0.02)} & \textbf{0.3571(0.02)} & \textbf{0.2181(0.03)} & \textbf{0.1227(0.06)} \\
    \midrule
    \multicolumn{3}{c}{\textbf{Average Improve (\%)}} 
& \multicolumn{1}{c}{$\downarrow2.1\%$} 
& \multicolumn{1}{c}{$\uparrow6.6\%$} 
& \multicolumn{1}{c}{$\uparrow0.5\%$} 
& \multicolumn{1}{c}{$\uparrow2.0\%$} 
& \multicolumn{1}{c}{$\uparrow5.2\%$} 
& \multicolumn{1}{c}{$\uparrow15.8\%$} \\

    \specialrule{1pt}{0pt}{0pt}  
  \end{tabular}} 
\end{table*}

\section{Empirical Evaluation}

We next evaluate the effect of this correction on validation-loss
estimation and influence ranking in multi-epoch training. We investigate the practical contribution of the repeated-deletion
correction identified in \autoref{sec:error-analysis}. Our experiments
evaluate three outcomes: the accuracy of validation-loss estimates, the
ranking of influential examples, and test performance after data
cleansing. We also examine estimation accuracy on noisy industrial data.
We compare estimated effects with leave-one-out retraining,
following established influence-evaluation
protocols~\citep{Data_cleaning,wang2025capturing,SOURCE}. For data cleansing,
we retrain on the selected subset and measure test
performance~\citep{Data_cleaning,SOURCE}. \appref{sec:rational} explains
these evaluation choices.

\subsection{Validation-Loss Change Estimation}
\label{subsec:exp_estimation}

\noindent\textbf{Evaluation task.}
Removing a training example changes both the learned parameters and the validation loss. We evaluate the loss change
$\Delta L_k^{\,i}=L(D_{\text{val}},\theta_k^i)-L(D_{\text{val}},\theta^i)$
using the linear influence
readout~\citep{koh2017understanding,hara2019data}. Each estimator supplies
a parameter displacement, which the readout maps to a loss change through
a first-order expansion of the validation loss. Comparing these estimates
with leave-one-out retraining measures loss-change accuracy and the
ability to identify the most influential examples.

\noindent\textbf{Experimental setup.}
We conduct 20 independent runs with different seeds on three binary
classification tasks: Adult~\citep{asuncion2007uci}, 20Newsgroups
(\texttt{ibm.pc.hardware} versus \texttt{mac.hardware}), and MNIST digits
$\{1,7\}$~\citep{726791}. We use a two-layer neural network for these
non-convex experiments~\citep{hara2019data}. Each run randomly samples 400 training and
400 validation instances.

We compare ACC-SGD-IE with SGD-IE against leave-one-out ground truth.
RMSE measures loss-change accuracy, Kendall's Tau measures agreement
with the ground-truth ranking, and Jaccard overlap measures agreement
among the top $\{10,30,50,70\}\%$ most influential examples. Controlled
feature and label noise~\citep{xia2022moderate} tests robustness to data
corruption. Results appear in \autoref{tab:all_influence} and
\autoref{fig:estimation_plot}; complementary convex controls appear in
\appref{app:results_convex}.

\noindent\textbf{Accuracy and ranking on clean data.}
ACC-SGD-IE reduces RMSE on all three benchmarks in Part~1 of
\autoref{tab:all_influence}: from $2.40\times10^{-3}$ to $2.20\times10^{-3}$
on Adult, from $1.82\times10^{-2}$ to $1.78\times10^{-2}$ on 20News, and
from $3.40\times10^{-3}$ to $2.00\times10^{-3}$ on MNIST. The mean relative
reduction is $17.24\%$. Kendall's Tau and Jaccard@10\% improve by $7.38\%$
and $7.66\%$ on average, respectively. The scatter plots in
\autoref{fig:estimation_plot} show closer agreement with leave-one-out
ground truth, particularly for high-influence examples. Together, these
results show that retaining the repeated-deletion interaction improves
both loss-change accuracy and influence ranking on the clean benchmarks.

\noindent\textbf{Robustness to feature noise.}
Part~2 of \autoref{tab:all_influence} evaluates five feature-noise settings. ACC-SGD-IE improves RMSE and Kendall's Tau in each setting. For Adult at $\sigma=0.01$, RMSE decreases from $3.80\times10^{-3}$ to $3.70\times10^{-3}$, while Kendall's Tau increases from $0.3016$ to $0.3352$. Across the five settings, the reported mean relative improvements are $17.22\%$ in RMSE, $38.46\%$ in Kendall's Tau, and $19.10\%$ in Jaccard@10\%. Mean Jaccard gains at the 30\%, 50\%, and 70\% cutoffs are $9.32\%$, $2.95\%$, and $1.52\%$, respectively. The correction therefore remains effective for aggregate accuracy and top-influence ranking under the tested input perturbations.

\noindent\textbf{Robustness to label noise.}
Part~3 of \autoref{tab:all_influence} flips a fraction $\rho\in\{0.1,0.3\}$ of training labels. ACC-SGD-IE reduces RMSE by $2.1\%$ on average, with the largest reported reduction of $4.4\%$ on 20News at $\rho=0.1$, and improves Kendall's Tau by $6.6\%$ on average. Mean Jaccard improvements are $15.8\%$, $5.2\%$, $2.0\%$, and $0.5\%$ at the 10\%, 30\%, 50\%, and 70\% cutoffs. The strongest ranking gains occur among the most influential examples, supporting the use of accumulative correction when identifying candidates for data cleansing.

\subsection{Cross-Epoch Influence Estimation Fidelity}
\label{subsec:cross-epoch-fidelity}

\begin{figure}[!t]
  \centering
  \includegraphics[width=\linewidth]{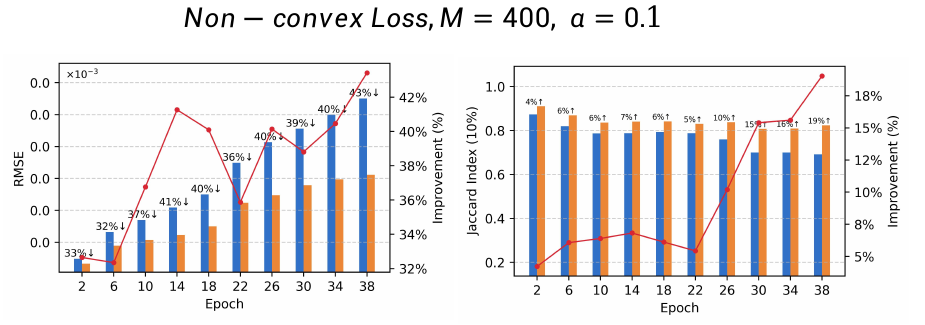}
  \caption{Cross-epoch estimation performance under non-convex training.
  Left: loss-change RMSE. Right: Jaccard@10\%. Blue bars denote SGD-IE,
  orange bars ACC-SGD-IE, and red curves the relative improvement.}
  \label{fig:degredation}
\end{figure}

When the target reappears, its deletion acts on the displacement
accumulated from earlier removals, with the resulting interaction
scaled by $\alpha/M$. We examine the contribution of this correction
across 40 training epochs using a neural network on 20News under
different batch sizes $M$ and learning rates $\alpha$. At epochs
$E\in\{2,6,\ldots,38\}$, we measure validation-loss change RMSE and
Jaccard@10\% against leave-one-out ground truth. Results are averaged
over 10 independent seeds. \autoref{fig:degredation} highlights configurations
with pronounced differences, and \autoref{fig:epoch-wise_both} reports
the complete set.

\noindent\textbf{Results across training epochs.}
Both estimators' errors increase with training in the displayed
non-convex configuration, with ACC-SGD-IE maintaining lower RMSE at
every displayed checkpoint. The relative RMSE reduction ranges from
$32\%$ to $43\%$. At epoch~38, SGD-IE's Jaccard@10\% is below $0.7$,
whereas ACC-SGD-IE remains above $0.8$, giving a $19\%$ relative ranking gain.

These trajectories show increasingly large benefits from accumulative
correction in the displayed configurations: ACC-SGD-IE preserves more
accurate loss-change estimates and more reliable top-influence rankings
over extended training. This behavior is consistent with the residual
analysis. The correction accounts for later removals acting on an
already displaced state, while the remaining Taylor error continues to
propagate. \appref{app:extended_cross_epoch} reports results across the
tested batch sizes and learning rates.

\subsection{Dataset Cleansing as a Downstream Validation}
\label{subsec:dataset_cleansing}

We next test whether the estimation and ranking gains improve data cleansing. Following~\citet{hara2019data}, we remove examples identified as harmful by their estimated influence and evaluate test misclassification after retraining.

\begin{figure}[!t]
  \centering
  \includegraphics[width=\linewidth]{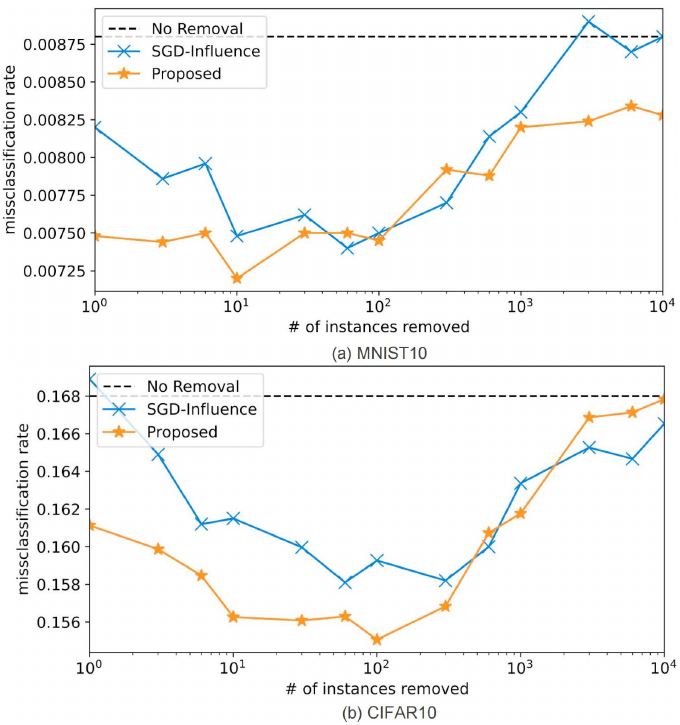}
  \caption{Data cleansing performance measured by mean test misclassification rate.}
  \label{fig:datacleansing}
\end{figure}

\noindent\textbf{Experimental setup.}
We evaluate on MNIST~\citep{726791} and CIFAR10~\citep{cifar10_100}, reserving 10,000 images from each training split for validation and using the remainder for SGD training. Two convolutional networks, one per dataset, train for 20 epochs with batch size 64 and initial learning rate 0.05, with all parameter updates checkpointed. After training, we compute per-example influence from the recorded checkpoints using SGD-IE and ACC-SGD-IE, remove the most-influential examples, and retrain from scratch under identical settings. We report test misclassification rate (MCR) averaged over five independent seeds in \autoref{fig:datacleansing}.

\noindent\textbf{Downstream benefits.}
ACC-SGD-IE improves test MCR at the highlighted removal budgets.
On MNIST, removing $m=10$ selected examples gives
approximately $0.72\%$ MCR, compared with approximately $0.75\%$ for SGD-IE. On CIFAR10,
ACC-SGD-IE reaches $15.5\%$ MCR at $m=100$, compared with approximately
$15.8\%$ for SGD-IE. These results connect the estimation gains to the
downstream task: more effective sample selection can improve test
performance after cleansing at these budgets.

\FloatBarrier
\section{Accuracy-Sensitive Industrial Applications}
\label{subsec:industrial_noise}

We use industrial fault and sensor datasets to test how accurately
the two methods estimate the effects of removing training samples.
We then examine their computational cost.

\subsection{Experiments on Industrial Data}
\label{subsec:industrial-accuracy}

We compare estimated parameter and validation-loss changes with
retraining results under controlled label noise.

\noindent\textbf{Experimental setup.}
We use Steel Plates Faults, Condition Monitoring of Hydraulic Systems,
and Gas Sensor Array Drift. For each dataset, we train with five
independent seeds. Each run uses 256 training samples, 128 validation
samples, and 128 uniformly sampled deletion targets. A neural network
with one hidden layer of eight tanh units trains for 40 epochs with batch size 32
and learning rate $0.1$. We change $10\%$ of training labels to a
different class and keep validation labels unchanged. We apply a fixed
global $\ell_2$ regularizer with coefficient $0.01$ to all parameters.

Steel and Hydraulic use stratified splits. We keep each Hydraulic cycle
as one observation and exclude derived efficiency channels.
Gas uses batches 1--6, 7--8, and 9--10 for training, validation, and
testing, respectively. Standardization is fitted on the selected
training rows; feature vectors are divided by the square root of
the feature count and rescaled to norm three if they exceed that norm.
The network starts with Xavier-normal weights and zero biases and uses
no batch normalization or dropout. Target sampling and label noise
use independent seeds. All computations use CPU float64.

For each target sample, we retrain after removing its data-loss
contribution at every occurrence. We keep the same initialization,
batch schedule, original batch denominator, and regularizer.
Both propagation operators include the penalty contribution
$-\alpha^i\rho_{\mathrm{reg}}I$, with $\rho_{\mathrm{reg}}=0.01$;
the deletion term $\Gamma_{k,i}$ is unchanged.

\noindent\textbf{Parameter and validation-loss accuracy.}
Parameter RMS error is the root mean square of the Euclidean errors
in the estimated final parameter changes across targets. We also
compute a plug-in loss change: we add each estimated parameter change
to the trained parameters and recompute the clean, class-balanced
validation negative log-likelihood. Plug-in RMSE compares these
nonlinear loss changes with those obtained by retraining.
For each metric, we divide the ACC-SGD-IE error by the SGD-IE error
within each seed and report the mean $\pm$ sample standard deviation
over five independent seeds. Ratios below one favor ACC-SGD-IE.

\begin{table}[!t]
 \centering
\caption{Estimation accuracy on noisy industrial data: ACC/SGD error ratios
(mean $\pm$ SD over five seeds).}
 \label{tab:recent-paired-faithfulness}
 \small
 \begin{tabular}{lcc}
  \toprule
  Dataset & Parameter RMS ratio & Plugin RMSE ratio \\
  \midrule
  Steel & $0.682\pm0.132$ & $0.830\pm0.105$ \\
  Hydraulic & $0.351\pm0.258$ & $0.565\pm0.314$ \\
  Gas drift & $0.364\pm0.221$ & $0.602\pm0.379$ \\
  \bottomrule
 \end{tabular}
\end{table}

\noindent\textbf{Results.}
\autoref{tab:recent-paired-faithfulness} reports lower mean parameter
and validation-loss errors on all three datasets. ACC-SGD-IE lowers
parameter RMS in all 15 runs and plug-in RMSE in 14 of 15 runs.

\noindent\textbf{Sensitivity to learning rate and label noise.}
Using the same tanh model, we test learning rates in
$\{0.05,0.1,0.2\}$ and label-change fractions over $\{0,0.2\}$.
The resulting 18 configurations use seed zero, with fixed target
pools and initializations. Parameter RMS decreases in all 18 settings,
with a mean relative reduction of $48.1\%$. Plug-in RMSE decreases
in 12 settings; the other six show increases of up to $49.7\%$.
Its mean relative reduction is $17.4\%$, weighting each configuration
equally. In these single-seed experiments, lower average parameter
error does not always give lower validation-loss error.

\noindent\textbf{Ablation of the repeated-deletion correction.}
We test the correction separately using linear classifiers on the same
three datasets, with the same preprocessing, $10\%$ label noise, and
regularizer. We use seed zero,
256 training samples, 128 validation samples, 256 test samples, 64 deletion targets,
80 epochs, batch size 32, and learning rate $0.1$. Steel and Gas use
softmax classifiers; Hydraulic uses binary logistic regression.
We remove the target's loss contribution at every appearance and vary
only when the correction is applied. The parameter change is zero
before the first removal, so correcting only then gives the same
estimate as SGD-IE. Correcting every other appearance
($1,3,5,\ldots$) reduces both errors on all three datasets.
Correcting all appearances after the first ($2,3,4,\ldots$) recovers
full ACC-SGD-IE and reduces parameter RMS by $45.7$--$75.5\%$
and plug-in RMSE by $48.5$--$61.9\%$ relative to SGD-IE.
This single-seed experiment tests the effect of applying the correction
at repeated removals.

\subsection{Computational Cost}
\label{subsec:industrial-computation}

\noindent\textbf{Computation and memory.}
Both forward SGD-IE and ACC-SGD-IE estimate parameter changes for
$Q$ samples over $N$ recorded SGD updates, requiring $\Theta(NQ)$
Hessian--vector products (HVPs). ACC-SGD-IE adds HVPs for a target
sample when it reappears. If only linear validation-loss change
scores are needed, SGD-IE can instead share one backward pass across
all targets, using $O(N)$ mini-batch HVPs plus gradient readouts at
target occurrences. This shared pass does not directly apply to
ACC-SGD-IE because each target uses a different operator $V_i^k$.

For a model with $p$ parameters, storing one vector for each of the
$Q$ samples requires $O(Qp)$ memory. Processing groups of $K$ samples
reduces the memory for these vectors to $O(Kp)$, plus storage for the
recorded training trajectory. Batched HVPs process several vectors
at once and can reduce runtime. The total work remains $\Theta(NQ)$.

\noindent\textbf{Runtime.}
We measure runtime on a Steel linear softmax classifier with an
intercept, using seed zero. We use 512 training samples, 256 validation
samples, and $10\%$ label noise. The global $\ell_2$ regularization
coefficient is $0.01$.
Training uses 160 epochs, batch size 16, and learning rate $0.1$,
giving 5,120 updates. Preprocessing follows
\autoref{subsec:industrial-accuracy}; the model and training settings
differ from the tanh accuracy experiments. We use CPU float64 with
one numerical thread and report the median of three wall-clock
measurements. Timings cover only estimation, excluding original
training, trajectory preparation, and retraining to obtain the true
effects.

\autoref{tab:recent-runtime} reports runtimes for different numbers
of queried samples. For 512 targets, forward SGD-IE takes $29.97$
seconds and forward ACC-SGD-IE takes $31.40$ seconds, an additional
$1.43$ seconds for the correction. Both methods return estimated parameter
changes and nonlinear plug-in loss changes, so they can be compared
directly. Shared backward SGD-IE takes $1.51$ seconds but returns
only linear scores. Its shorter runtime shows the benefit of sharing
the calculation when only linear scores are needed.

\begin{table}[!htbp]
\centering
\caption{Runtime on Steel (seconds).}
\label{tab:recent-runtime}
\small
\setlength{\tabcolsep}{4pt}
\begin{tabular}{@{}rrrr@{}}
\toprule
Samples & \makecell{SGD-IE\\Backward} & \makecell{SGD-IE\\Forward} & \makecell{ACC-SGD-IE\\Forward} \\
\midrule
1 & $2.056$ & $1.194$ & $1.511$ \\
8 & $1.615$ & $1.707$ & $1.887$ \\
64 & $1.373$ & $4.324$ & $5.405$ \\
512 & $1.505$ & $29.967$ & $31.400$ \\
\bottomrule
\end{tabular}
\end{table}
\FloatBarrier
\section{Limitations and Discussion}
\label{sec:limitations-discussion}

The experiments cover a limited set of datasets, models, and training
settings. They do not establish that the same improvements hold for
other tasks, model architectures, or applications.

The current time and memory requirements may limit use on larger
models and datasets. Practical performance at the scale of large
deep learning models and LLM applications remains untested.

\section{Conclusions and Future Work}

We identify an interaction that standard multi-epoch SGD-IE omits
when estimating the effect of removing a sample throughout training.
ACC-SGD-IE adds the corresponding correction. The analysis bounds
the remaining Taylor error and shows that the estimate is exact for
quadratic losses under the fixed training protocol. Experiments show
lower average estimation errors and gains in ranking and data
cleansing in the tested settings.

Future work will focus on improving scalability and adapting the
method to modern deep learning and LLM applications. We will evaluate
its accuracy, runtime, and memory use on larger models and datasets.

\FloatBarrier
\section*{Acknowledgments}
OpenAI Codex was used to improve the writing and presentation
throughout the manuscript. ChatGPT (OpenAI) assisted in exploring
the ACC-SGD-IE error bounds reported in Section~\ref{sec:error-analysis}
and the supporting analysis in \appref{app:error-bounds}.
The research ideas, experimental design and code, original figures
and table data, and identification of the scientific limitations
were developed by the authors without AI assistance.
The authors are responsible for the final manuscript.

\clearpage
\appendices

\section*{APPENDIX}
\autoref{tab:notations} collects the symbols used in the main text and proofs.
The appendices provide the estimator derivations and error analysis,
followed by implementation details and supplementary evaluations.
The following guide summarizes the content of each appendix:
\begin{itemize}[leftmargin=*,itemsep=2pt]
\item \appref{sec:acc_derivation} derives and unrolls the ACC-SGD-IE
recurrence from the coupled SGD updates, and proves the sample-weight
derivative interpretation of SGD-IE.
\item \appref{app:bias} uses a two-occurrence example to illustrate how
a later removal interacts with the displacement created by an earlier
removal.
\item \appref{app:error-recursions} proves the two exact error recursions
and separates the deleted-sample Hessian interaction from the retained-data
Taylor remainder.
\item \appref{app:error-bounds} proves finite-horizon bounds for non-convex
and strongly convex objectives, and compares the resulting error envelopes.
\item \appref{sec:acc_implementation} distinguishes full-vector and scalar-query
costs, and describes target sharding, batched HVPs, and the scope of
implementation approximations.
\item \appref{app:additional-experiments} gives evaluation protocols,
convex-model loss-change results, and extended cross-epoch comparisons
across batch sizes and learning rates.
\item \appref{sec:extension} evaluates accumulative correction in DVEmb
and Adam-IE through ranking comparisons on MNIST.
\end{itemize}

\begin{table*}[!t]
\centering
\caption{Comprehensive summary of notation used throughout this work.}
\begin{tabular}{@{}p{0.16\textwidth}P{0.78\textwidth}@{}}
\toprule
\textbf{Symbol} & \textbf{Definition} \\
\midrule
$D $ 
  & Training dataset of $n$ samples. \\
$n$ 
  & Total number of training samples. \\
$x$ 
  & Input feature vector of dimension $d$. \\
$y$ 
  & One-hot encoded label vector over $c$ classes. \\
$p$ 
  & Dimensionality of the parameter vector $\theta\in\mathbb{R}^p$. \\
$\theta$ 
  & Model parameter vector. \\
$f(x;\theta)$ 
  & Model’s pre-softmax output for input $x$. \\
$\ell(\cdot,\cdot)$ 
  & Pointwise loss function (e.g.\ cross‐entropy). \\
$L(z,\theta)$ 
  & Loss incurred by sample $z=(x,y)$. \\
$L(D,\theta)$ 
  & Full-dataset loss. \\
$g(z,\theta)$ 
  & Gradient of the loss w.r.t.\ $\theta$. \\
$g(D,\theta)$ 
  & Full-dataset gradient. \\
$H(z,\theta)$ 
  & Hessian of the loss w.r.t.\ $\theta$. \\
$H(D,\theta)$ 
  & Full-dataset Hessian. \\
$\theta^{0}$ 
  & Initial parameter vector before any SGD updates. \\
$T$ 
  & Total number of epochs. \\
$B$ 
  & Number of mini-batches per epoch. \\
$N$ 
  & Total number of SGD updates. \\
$Z^i$ 
  & Mini-batch for the update from $\theta^i$ to $\theta^{i+1}$, $0\le i<N$; $M=|Z^i|$ in the fixed-size analysis. \\
$M$ 
  & Mini-batch size. \\
$\alpha^i$ 
  & Learning rate at iteration $i$. \\
$\theta^i$ 
  & Parameter vector after $i$ SGD updates, $0\le i\le N$. \\
$\theta_k^i$ 
  & Parameter vector after $i$ updates when $z_k$ is omitted (Counterfactual SGD). \\
$\Delta\theta_k^i$ 
  & True influence of sample $z_k$ at iteration $i$. \\
$\widehat{\Delta\theta_k^i}$ 
  & SGD-IE estimated influence of sample $z_k$ at iteration $i$. \\
$\widetilde{\Delta\theta_k^i}$ 
  & ACC-SGD-IE estimated influence of sample $z_k$ at iteration $i$. \\
$\pi_t(k)$ 
  & Zero-based update/batch index of the $t$-th occurrence of $z_k$, $0\le\pi_t(k)<N$. \\
$U_i $ 
  & Defined by \(U_{i}:=I-\alpha^{i}H(Z^{i},\theta^{i})\) in \autoref{subsec:SGD-IE-Definition}. \\
$\Phi_{j,i}$ 
  & Defined by $\Phi_{j,i}
  := U_j U_{j-1}\cdots U_i,\quad (j\ge i)$; an empty product is $I$, as in \autoref{eq:error-sum-classic}. \\
$\widetilde\Phi_{j,i}^k$
  & Ordered ACC transition product $V_j^kV_{j-1}^k\cdots V_i^k$; an empty product is $I$. \\
$\iota_i^k$
  & Occurrence indicator $\mathbb{I}(z_k\in Z^i)$. \\
$c_i^k$
  & Retained batch fraction $1-\iota_i^k/M$ in the fixed-size analysis. \\
$V_i^k$ 
  & $V_i^k 
  = U_i 
    + \mathbb{I}(z_k\in Z^i)\,
      \frac{\alpha^i}{|Z^i|}\,
      H\bigl(z_k,\theta^i\bigr)$, defined in \autoref{eq:define_V}. \\
$\Gamma_{k,i}$ 
  & $\Gamma_{k,i}
  = \mathbb{I}(z_k\in Z^i)\,
      \frac{\alpha^i}{|Z^i|}\,
      g\bigl(z_k,\theta^i\bigr)$, defined in \autoref{eq:define_Gamma}. \\
$E_k^i$ 
  & Estimation error of SGD-IE. \\
$\widetilde E_k^i$ 
  & Estimation error of ACC-SGD-IE. \\
$R_k^i$ 
  & SGD-IE residual in \autoref{eq:error-recursion-classic}: $R_k^i=\widetilde R_k^i-\frac{\iota_i^k}{|Z^i|}H(z_k,\theta^i)\Delta\theta_k^i$. \\
$\widetilde R_k^i$ 
  & Retained-data gradient Taylor residual in \autoref{eq:acc-residual}, averaged with the original denominator $|Z^i|$ and evaluated using the true displacement $\Delta\theta_k^i$. \\
$A_i$
  & Cumulative learning-rate budget before update $i$: $A_i=\sum_{j=0}^{i-1}\alpha^j$, with $A_0=0$. \\
$G$
& Bound on the removed-record gradient at original occurrence states: $\|g(z_k,\theta^i)\|_{2}\le G$ when $\iota_i^k=1$. \\
$L$ 
  & $L$-Lipschitz Hessians: $\|\nabla^2L(z,\theta_1)-\nabla^2L(z,\theta_2)\|\le L\|\theta_1-\theta_2\|$. \\
$\Lambda$
& Uniform spectral bound $\|H(z,\theta)\|_{2}\le\Lambda$. \\

\bottomrule
\end{tabular}

\label{tab:notations}
\end{table*}

\section{Deriving the ACC-SGD-IE Recurrence}
\label{sec:acc_derivation}

We derive the recurrence from the coupled, fixed-denominator updates in
\autoref{eq:Counterfactual_SGD}: record
identities, batches, learning rates and stochastic choices are shared,
and each record appears at most once in a batch. Matrix products place
the latest update on the left.

\subsection{Subtracting the Two SGD Updates}

We recall the ordinary SGD update (\autoref{eq:SGD})
\[
  \theta^{i+1}
  \;=\;
  \theta^{i}
  \;-\;
  \frac{\alpha^{\,i}}{\lvert Z^{\,i}\rvert}
  \sum_{z\in Z^{\,i}} g\!\bigl(z,\theta^{\,i}\bigr),
\]
and the counterfactual update with $z_k$ removed (\autoref{eq:Counterfactual_SGD})
\[
  \theta_k^{\,i+1}
  \;=\;
  \theta_k^{\,i}
  \;-\;
  \frac{\alpha^{\,i}}{\lvert Z^{\,i}\rvert}
  \sum_{z\in Z^{\,i}\setminus\{z_k\}}
    g\!\bigl(z,\theta_k^{\,i}\bigr).
\]
Defining the influence
\(
  \Delta\theta_k^{\,i} := \theta_k^{\,i} - \theta^{\,i},
\)
we subtract the two updates to obtain (\autoref{eq:model_change})
\begin{equation}\label{eq:app_model_change}
\begin{split}
\Delta\theta_k^{\,i+1}
&=
\Delta\theta_k^{\,i}
+\frac{\alpha^{\,i}}{\lvert Z^{\,i}\rvert}
\Biggl(
  \sum_{z\in Z^{\,i}} g(z,\theta^{\,i}) \\
&\qquad\;-\;
  \sum_{z\in Z^{\,i}\setminus\{z_k\}}
    g\bigl(z,\theta_k^{\,i}\bigr)
\Biggr).
\end{split}
\end{equation}

\subsection{Before, At, and After the First Occurrence}

For $T$ complete epochs with one appearance per record in each epoch, let
\(\pi_1(k)<\dots<\pi_T(k)\) be the indices of the \(T\) mini-batches
containing \(z_k\). For each update step $i=0,\ldots,N-1$, three situations arise:
\begin{enumerate}[label=(\alph*)]
  \item \textbf{Pre-occurrence} (\(i<\pi_1(k)\)): Since \(z_k\notin Z^i\) and \(\theta_k^{\,i}=\theta^{\,i}\),
        both sums in~\autoref{eq:app_model_change} cancel, yielding
        \(\Delta\theta_k^{\,i+1}=0\).

  \item \textbf{First appearance} (\(i=\pi_1(k)\)):
        Here \(Z^{\,i}\setminus\{z_k\}=Z^{\,i}_k\), and since
        \(\Delta\theta_k^{\,i}=0\), the difference of sums reduces exactly to
        \(\;g(z_k,\theta^{\,\pi_1(k)})\), giving
        \[
          \Delta\theta_k^{\,i+1}
          =
          \frac{\alpha^{\,\pi_1(k)}}{\lvert Z^{\,\pi_1(k)}\rvert}\,
          g\!\bigl(z_k,\theta^{\,\pi_1(k)}\bigr).
        \]

  \item \textbf{Propagation} (\(i>\pi_1(k)\)):
        \begin{itemize}[leftmargin=0pt]
          \item If \(z_k\notin Z^{\,i}\), then
                \(\theta_k^{\,i}=\theta^{\,i}+\Delta\theta_k^{\,i}\)
                enters only through the gradient argument. Write the exact
                first-order Taylor decomposition as
                \[
                g(z,\theta_k^{\,i})
                =
                g(z,\theta^{\,i})
                + H(z,\theta^{\,i})\,\Delta\theta_k^{\,i}
                + r_z^i,
                \]
                where
                $r_z^i:=g(z,\theta_k^i)-g(z,\theta^i)
                -H(z,\theta^i)\Delta\theta_k^i$.
                This yields
                \[
                \Delta\theta_k^{\,i+1}
                =
                U_i\,\Delta\theta_k^{\,i}
                -\alpha^i\widetilde R_k^i,
                \]
                where \(U_i=I-\alpha^iH(Z^i,\theta^i)\) and
                $\widetilde R_k^i=|Z^i|^{-1}
                \sum_{z\in Z^i\setminus\{z_k\}}r_z^i$.
                Dropping this residual gives the homogeneous linearized update.

          \item If \(z_k\in Z^{\,i}\), then
                \[
                \begin{aligned}
                \Delta\theta_k^{\,i+1}
                &=
                \Delta\theta_k^{\,i}
                +\frac{\alpha^{\,i}}{\lvert Z^{\,i}\rvert}
                \Biggl[
                g(z_k,\theta^i)
                \\
                &\qquad +
                \sum_{z\in Z^i\setminus\{z_k\}}
                \Bigl(
                g(z,\theta^i)-g(z,\theta_k^i)
                \Bigr)
                \Biggr].
                \end{aligned}
                \]
                Applying the same exact Taylor decomposition to the remaining
                samples gives
                \[
                \begin{aligned}
                \Delta\theta_k^{\,i+1}
                \;=\; &
                U_i\,\Delta\theta_k^{\,i}
                \\
                &\;+\;
                \frac{\alpha^{\,i}}{\lvert Z^{\,i}\rvert}
                \Bigl[
                  H\bigl(z_k,\theta^{\,i}\bigr)\,\Delta\theta_k^{\,i}
                  + g\bigl(z_k,\theta^{\,i}\bigr)
                \Bigr]
                \\
                &-\alpha^i\widetilde R_k^i.
                \end{aligned}
                \]
        \end{itemize}
\end{enumerate}

\subsection{Unified Recurrence}
\label{app:unified-derivation}

The indicator \(\mathbb{I}(z_k\in Z^i)\) combines the cases where the
target is absent or present. The resulting exact recurrence is
\autoref{eq:recursive_formula}:
\begin{equation}\label{eq:app_recursive_formula}
  \Delta\theta_k^{\,i+1}
  \;=\;
  V_i^k\,\Delta\theta_k^{\,i}
  \;+\;
  \Gamma_{k,i}
  \;-\;\alpha^i\widetilde R_k^i,
\end{equation}
where (\autoref{eq:define_V}--\autoref{eq:define_Gamma})
\begin{align}
  V_i^k
    &=
    U_i
    \;+\;
    \mathbb{I}\!\bigl(z_k\in Z^i\bigr)
    \frac{\alpha^i}{\lvert Z^i\rvert}\,
    H\bigl(z_k,\theta^i\bigr),
    \label{eq:app_define_V}\\
  \Gamma_{k,i}
    &=
    \mathbb{I}\!\bigl(z_k\in Z^i\bigr)
    \frac{\alpha^{\,i}}{\lvert Z^{\,i}\rvert}\,
    g\!\bigl(z_k,\theta^{\,i}\bigr).
    \label{eq:app_define_Gamma}
\end{align}
The retained-sample residual is
\[
  \widetilde R_k^i
  =\frac1{|Z^i|}\sum_{z\in Z^i\setminus\{z_k\}}
   \bigl[g(z,\theta_k^i)-g(z,\theta^i)
       -H(z,\theta^i)\Delta\theta_k^i\bigr].
\]
If the retained-sample Hessians are $L$-Lipschitz on the connecting
segments, the integral Taylor formula yields
\[
  \begin{aligned}
  r_z^i=\int_0^1
   \bigl[H(z,\theta^i+u\Delta\theta_k^i)-H(z,\theta^i)\bigr]
   \Delta\theta_k^i\,du,\\
  \|r_z^i\|_2\le\frac L2\|\Delta\theta_k^i\|_2^2.
  \end{aligned}
\]
Consequently,
\[
  \|\widetilde R_k^i\|_2
  \le\frac{|Z^i|-\mathbb I(z_k\in Z^i)}{|Z^i|}
       \frac L2\|\Delta\theta_k^i\|_2^2.
\]
The occurrence-level correction is explained in
\autoref{sec:Accumulative SGD-Influence}.

\subsection{Unrolling the Recurrence}
\label{app:closed-form}

Dropping $\widetilde R_k^i$ defines
$\widetilde{\Delta\theta}_k^{i+1}
=V_i^k\widetilde{\Delta\theta}_k^i+\Gamma_{k,i}$.
Starting from $\widetilde{\Delta\theta}_k^0=0$, repeated substitution
gives the contribution from each occurrence, propagated through all
later updates. Induction on $i$ yields
\begin{equation}\label{eq:app_acc_unroll}
  \widetilde{\Delta\theta}_k^{\,i}
  \;=\;
  \sum_{t:\,\pi_t(k)<i}
    \Bigl(
      V_{i-1}^k
      V_{i-2}^k
      \cdots
      V_{\pi_t(k)+1}^k
    \Bigr)
    \Gamma_{k,\pi_t(k)},
\end{equation}
where an empty product is the identity. This is the estimator in
\autoref{eq:estimator}.

\subsection{SGD-IE as a Sample-Weight Derivative}
\label{app:weight-tangent-proof}

We prove \autoref{prop:sgdie-weight-tangent} by varying the removal weight
continuously from zero to one and differentiating the coupled updates.
\begin{proof}
The interpolated update is
\[
  \begin{aligned}
  \theta_k^{i+1}(\varepsilon)
  &=\theta_k^i(\varepsilon)
   -\frac{\alpha^i}{|Z^i|}
    \left[\sum_{z\in Z^i}g(z,\theta_k^i(\varepsilon))
     \right.\\
  &\qquad\left.
     -\varepsilon\mathbb I(z_k\in Z^i)
        g(z_k,\theta_k^i(\varepsilon))\right].
  \end{aligned}
\]
Its endpoints are $\theta_k^i(0)=\theta^i$ and
$\theta_k^i(1)=\theta_k^i$. Differentiating at zero gives
\[
  \left.\frac{d\theta_k^{i+1}(\varepsilon)}{d\varepsilon}\right|_0
  =U_i\left.\frac{d\theta_k^i(\varepsilon)}{d\varepsilon}\right|_0
   +\Gamma_{k,i}.
\]
The term differentiating the removed gradient through
$\theta_k^i(\varepsilon)$ carries a factor $\varepsilon$ and vanishes at
zero. The initialization has zero derivative. This is exactly the
recurrence whose unrolling gives
\autoref{eq:SGD_Influence_Estimation}, so induction proves the tangent
identity. If the path is continuously differentiable on $[0,1]$, its
finite displacement satisfies
\[
  \Delta\theta_k^i
  =\int_0^1\frac{d\theta_k^i(\varepsilon)}{d\varepsilon}\,d\varepsilon,
\]
which need not equal the derivative evaluated at zero.
\end{proof}

The correction has different orders under these two descriptions. It
is linear in the finite displacement, whereas along the weight
interpolation it is proportional to
$\varepsilon H(z_k,\theta^i)
[\theta_k^i(\varepsilon)-\theta^i]$ and is $O(\varepsilon^2)$ when the
displacement is $O(\varepsilon)$. ACC-SGD-IE retains this interaction; a
complete second-order expansion in $\varepsilon$ would also include
the nonlinear Taylor terms of the retained gradients.

\section{Case Study: Two-Occurrence Illustration of the Deletion Interaction}
\label{app:bias}

The trajectory in \autoref{fig:framework} has five updates,
$Z^0,Z^1,Z^2,Z^3,Z^4$, with target occurrences
$\pi_1(k)=1$ and $\pi_2(k)=3$. With the notation of
\autoref{sec:Accumulative SGD-Influence}, SGD-IE gives
\[
 \widehat{\Delta\theta_k^5}
 =U_4U_3U_2\Gamma_{k,1}+U_4\Gamma_{k,3}.
\]
We compare its second contribution with the effect of removing the
sample after the first removal has already changed the trajectory.

\paragraph{Conditional second removal.}
Let $\theta_k^i[\pi_1(k)]$ remove $z_k$ only at its first occurrence,
keeping the recorded batches and original denominators. The exact
repeated-removal displacement decomposes as
\[
 \Delta\theta_k^5
 =\underbrace{\theta_k^5[\pi_1(k)]-\theta^5}_{\text{first-removal effect}}
 +\underbrace{\theta_k^5-\theta_k^5[\pi_1(k)]}_{\text{conditional second-removal effect}}.
\]
The repeated-removal and first-removal-only trajectories coincide before
update $3$. Removing $z_k$ there therefore gives the exact difference
\[
 \theta_k^4-\theta_k^4[\pi_1(k)]
 =\frac{\alpha^3}{|Z^3|}
   g\bigl(z_k,\theta_k^3[\pi_1(k)]\bigr).
\]
Since $z_k\notin Z^4$, a first-order expansion of the next update gives
\[
 \begin{aligned}
 \theta_k^5-\theta_k^5[\pi_1(k)]
 &\approx
 \bigl[I-\alpha^4H(Z^4,\theta_k^4[\pi_1(k)])\bigr]
 \\
 &\qquad
 \frac{\alpha^3}{|Z^3|}
 g\bigl(z_k,\theta_k^3[\pi_1(k)]\bigr).
 \end{aligned}
\]
SGD-IE instead uses the observed states in its contribution
\[
 U_4\Gamma_{k,3}
 =\bigl[I-\alpha^4H(Z^4,\theta^4)\bigr]
   \frac{\alpha^3}{|Z^3|}g(z_k,\theta^3).
\]
The conditional removal thus acts at a state changed by the first
removal, whereas this independently injected contribution is evaluated
at the observed state.

\paragraph{The term retained by ACC-SGD-IE.}
The corrected estimate is
\[
 \begin{aligned}
 \widetilde{\Delta\theta}_k^5
 &=U_4V_3^kU_2\Gamma_{k,1}+U_4\Gamma_{k,3},\\
 V_3^k&=U_3+\frac{\alpha^3}{|Z^3|}H(z_k,\theta^3).
 \end{aligned}
\]
Subtracting the SGD-IE estimate isolates the correction:
\[
 \widetilde{\Delta\theta}_k^5-\widehat{\Delta\theta_k^5}
 =U_4\frac{\alpha^3}{|Z^3|}H(z_k,\theta^3)U_2\Gamma_{k,1}.
\]
The sample Hessian acts on the displacement propagated from the first
removal. ACC-SGD-IE retains this interaction while linearizing the
retained gradients; \appref{app:error-recursions} derives the remaining
residual.

\section{Exact Error Recursions and Residual Cancellation}
\label{app:error-recursions}

Both estimators approximate the same fixed-denominator counterfactual
displacement. Subtracting that displacement from each estimator gives
an exact error recurrence: earlier errors propagate through an update
matrix, and each step adds a local residual. We derive the SGD-IE and
ACC-SGD-IE recurrences separately.

\subsection{SGD-IE: Recursion and Proof}
\label{app:sgdie-error-proof}

\begin{proposition}[\sgdiehl{Estimation Error of SGD-IE}]
\label{prop:sgdie-error-recursion}
Let $\widehat{\Delta\theta_k^i}$ denote the estimation of the SGD-Influence Estimator, and define
\[
  E_k^i
  :=
  \widehat{\Delta\theta_k^i}-\Delta\theta_k^i
\]
as the estimation error of SGD--IE at iteration \(i\) for sample \(z_k\). With
\[
  U_i
  :=
  I-\alpha^i H(Z^i,\theta^i),
\]
defined in \autoref{eq:One_Epoch_SGD_Influence_Estimation}, the estimation error satisfies the recursion
\begin{equation}\label{eq:error-recursion-classic}
\begin{aligned}
  E_k^{\,i+1}
  &=
  U_i E_k^i
  +
  \alpha^i R_k^i,\\
  R_k^i
  &:=
  \frac{1}{|Z^i|}
  \sum_{z\in Z^i\setminus\{z_k\}}
  \Bigl[
    g(z,\theta_k^i)-g(z,\theta^i)
  \Bigr] \\
  &\quad -
  H(Z^i,\theta^i)\Delta\theta_k^i .
\end{aligned}
\end{equation}
Equivalently, the residual has the exact split
\begin{equation}\label{eq:residual-split}
  R_k^i=\widetilde R_k^i
   -\mathbb I(z_k\in Z^i)\frac1{|Z^i|}
      H(z_k,\theta^i)\Delta\theta_k^i,
\end{equation}
where $\widetilde R_k^i$ is the retained-sample Taylor residual in
\autoref{eq:acc-residual}.
Moreover, unrolling this recursion from \(i=0\) to \(N-1\) yields
\begin{equation}\label{eq:error-sum-classic}
\sgdiehl{
\begin{aligned}
E_k^N
  &=
  \sum_{i=0}^{N-1}
    \alpha^i
    \Phi_{N-1,i+1}
    R_k^i, \\
  \Phi_{j,i}
  &:=
  U_j U_{j-1}\cdots U_i,
  \quad
  (j\ge i),
\end{aligned}
}
\end{equation}
where by convention \(\Phi_{i,i+1}=I\).
\end{proposition}

\begin{proof}
\textit{Step 1: Subtract the true displacement update from the estimator.}
The true influence satisfies
\[
\begin{aligned}
\Delta\theta_k^{\,i+1}
&=
\Delta\theta_k^{\,i} \\
&\quad -
\alpha^{\,i}
\frac{1}{|Z^i|}
\sum_{z\in Z^i\setminus\{z_k\}}
\bigl[
  g(z,\theta_k^{\,i})-g(z,\theta^{\,i})
\bigr] \\
&\quad +
\mathbb{I}(z_k\in Z^i)
\frac{\alpha^i}{|Z^i|}
g(z_k,\theta^i).
\end{aligned}
\]
The multi-epoch SGD-IE estimator in \autoref{eq:SGD_Influence_Estimation} can be equivalently written in the recursive form
\[
\widehat{\Delta\theta_k}^{\,i+1}
=
U_i\widehat{\Delta\theta_k^{\,i}}
+
\mathbb{I}(z_k\in Z^i)
\frac{\alpha^i}{|Z^i|}
g(z_k,\theta^i).
\]
Each occurrence injects a gradient contribution that is propagated
through subsequent $U_i$ updates. This is the sample-weight derivative
proved in \appref{app:weight-tangent-proof}.

By the error definition
\[
  E_k^{\,i}
  =
  \widehat{\Delta\theta_k^{\,i}}
  -
  \Delta\theta_k^{\,i},
\]
we have
\[
\begin{aligned}
E_k^{\,i+1}
&=
\widehat{\Delta\theta_k}^{\,i+1}
-
\Delta\theta_k^{\,i+1} \\
&=
U_i\widehat{\Delta\theta_k^{\,i}}
+
\mathbb{I}(z_k\in Z^i)
\frac{\alpha^i}{|Z^i|}
g(z_k,\theta^i) \\
&\quad -
\left[
\begin{aligned}
&\Delta\theta_k^{\,i} \\
&-
\alpha^{\,i}
\frac{1}{|Z^i|}
\sum_{z\in Z^i\setminus\{z_k\}}
\bigl[
  g(z,\theta_k^{\,i})-g(z,\theta^{\,i})
\bigr] \\
&+
\mathbb{I}(z_k\in Z^i)
\frac{\alpha^i}{|Z^i|}
g(z_k,\theta^i)
\end{aligned}
\right].
\end{aligned}
\]
\textit{Step 2: Isolate the residual.}
The occurrence-level injection terms cancel. Insert
\[
  \widehat{\Delta\theta_k^{\,i}}
  =
  E_k^{\,i}
  +
  \Delta\theta_k^{\,i},
\]
and use
\[
  I-U_i
  =
  \alpha^i H(Z^i,\theta^i).
\]
Then
\[
\begin{aligned}
E_k^{\,i+1}
&=
U_i
\bigl(
E_k^{\,i}+\Delta\theta_k^{\,i}
\bigr)
-
\Delta\theta_k^{\,i} \\
&\quad +
\alpha^{\,i}
\frac{1}{|Z^i|}
\sum_{z\in Z^i\setminus\{z_k\}}
\bigl[
  g(z,\theta_k^{\,i})-g(z,\theta^{\,i})
\bigr] \\
&=
U_iE_k^{\,i}
-
\alpha^i H(Z^i,\theta^i)\Delta\theta_k^i \\
&\quad +
\alpha^{\,i}
\frac{1}{|Z^i|}
\sum_{z\in Z^i\setminus\{z_k\}}
\bigl[
  g(z,\theta_k^{\,i})-g(z,\theta^{\,i})
\bigr] \\
&=
U_iE_k^{\,i} \\
&\quad +
\alpha^i
\left\{
\begin{aligned}
&\frac{1}{|Z^i|}
\sum_{z\in Z^i\setminus\{z_k\}}
\bigl[
  g(z,\theta_k^{\,i})-g(z,\theta^{\,i})
\bigr] \\
&-
H(Z^i,\theta^i)\Delta\theta_k^i
\end{aligned}
\right\} \\
&=
U_iE_k^{\,i}
+
\alpha^i R_k^i .
\end{aligned}
\]
\textit{Step 3: Unroll the error and separate its two sources.}
The preceding calculation proves \autoref{eq:error-recursion-classic}.
Unrolling the recursion with
\[
  \Phi_{j,i}
  =
  U_jU_{j-1}\cdots U_i
\]
gives \autoref{eq:error-sum-classic}.
Subtracting the retained-sample Hessian sum from the full mini-batch
Hessian also gives
\[
 \begin{aligned}
 &\frac1{|Z^i|}\sum_{z\in Z^i\setminus\{z_k\}}H(z,\theta^i)
 -H(Z^i,\theta^i) \\
 &\qquad =-\mathbb I(z_k\in Z^i)\frac1{|Z^i|}H(z_k,\theta^i),
 \end{aligned}
\]
which proves the residual split in \autoref{eq:residual-split}.
\end{proof}

\subsection{ACC-SGD-IE: Recursion and Proof}
\label{app:acc-error-proof}

\begin{proposition}[\acchl{Estimation Error of ACC-SGD-IE}]
\label{prop:acc-error-recursion}
Let $\widetilde{\Delta\theta}_k^i$ denote the ACC-SGD-IE estimate, and define
\[
  \widetilde E_k^i
  :=
  \widetilde{\Delta\theta}_k^i-\Delta\theta_k^i .
\]
Let
\[
  V_i^k
  :=
  U_i
  +
  \mathbb{I}(z_k\in Z^i)
  \frac{\alpha^i}{|Z^i|}
  H(z_k,\theta^i),
\]
where
\[
  U_i
  :=
  I-\alpha^i H(Z^i,\theta^i).
\]
Then the ACC-SGD-IE estimation error satisfies
\begin{equation}\label{eq:acc-error-rec}
  \widetilde E_k^{\,i+1}
  =
  V_i^k \widetilde E_k^i
  +
  \alpha^i \widetilde R_k^i,
\end{equation}
where
\begin{equation}\label{eq:acc-residual}
  \widetilde R_k^i
  :=
  \frac{1}{|Z^i|}
  \sum_{z\in Z^i\setminus\{z_k\}}
  \left[
  \begin{aligned}
    &g(z,\theta_k^i)-g(z,\theta^i) \\
    &-
    H(z,\theta^i)\Delta\theta_k^i
  \end{aligned}
  \right].
\end{equation}
Consequently,
\begin{equation}\label{eq:error-sum-acc}
\acchl{
  \begin{aligned}
  \widetilde E_k^N
  &=
  \sum_{i=0}^{N-1}
  \alpha^i
  \widetilde\Phi_{N-1,i+1}^k
  \widetilde R_k^i, \\
  \widetilde\Phi_{j,i}^k
  &:=
  V_j^k V_{j-1}^k\cdots V_i^k,
  \quad
  (j\ge i),
  \end{aligned}
}
\end{equation}
where by convention \(\widetilde\Phi_{i,i+1}^k=I\).
\end{proposition}

\begin{proof}
\textit{Step 1: Express the update using the retained-batch Hessian.}
The ACC-SGD-IE transition matrix is
\[
  V_i^k
  =
  U_i
  +
  \mathbb{I}(z_k\in Z^i)
  \frac{\alpha^i}{|Z^i|}
  H(z_k,\theta^i).
\]
Equivalently, since
\[
  U_i
  =
  I-\alpha^i H(Z^i,\theta^i),
\]
and \(H(Z^i,\theta^i)\) denotes the mini-batch averaged Hessian, we have
\[
  V_i^k
  =
  I
  -
  \alpha^i
  \frac{1}{|Z^i|}
  \sum_{z\in Z^i\setminus\{z_k\}}
  H(z,\theta^i).
\]
The ACC-SGD-IE estimator recursion can therefore be written as
\[
\widetilde{\Delta\theta}_k^{\,i+1}
=
V_i^k
\widetilde{\Delta\theta}_k^{\,i}
+
\alpha^i
\frac{1}{|Z^i|}
\mathbb{I}(z_k\in Z^i)
g(z_k,\theta^i).
\]
\textit{Step 2: Separate the retained-sample Taylor remainder.}
The true influence recursion is
\[
\begin{aligned}
\Delta\theta_k^{\,i+1}
&=
\Delta\theta_k^{\,i} \\
&\quad -
\alpha^i
\frac{1}{|Z^i|}
\sum_{z\in Z^i\setminus\{z_k\}}
\bigl[
  g(z,\theta_k^i)-g(z,\theta^i)
\bigr] \\
&\quad +
\alpha^i
\frac{1}{|Z^i|}
\mathbb{I}(z_k\in Z^i)
g(z_k,\theta^i).
\end{aligned}
\]
Using the definition of \(V_i^k\), we can rewrite the preceding recursion as
\[
\Delta\theta_k^{\,i+1}
=
V_i^k\Delta\theta_k^i
+
\alpha^i
\frac{1}{|Z^i|}
\mathbb{I}(z_k\in Z^i)
g(z_k,\theta^i)
-
\alpha^i \widetilde R_k^i,
\]
where
\[
  \widetilde R_k^i
  =
  \frac{1}{|Z^i|}
  \sum_{z\in Z^i\setminus\{z_k\}}
  \left[
  \begin{aligned}
    &g(z,\theta_k^i)-g(z,\theta^i) \\
    &-
    H(z,\theta^i)\Delta\theta_k^i
  \end{aligned}
  \right].
\]
\textit{Step 3: Subtract and unroll.}
Subtracting the true recursion from the estimator recursion yields
\[
\begin{aligned}
\widetilde E_k^{\,i+1}
&=
\widetilde{\Delta\theta}_k^{\,i+1}
-
\Delta\theta_k^{\,i+1} \\
&=
V_i^k
\bigl(
\widetilde{\Delta\theta}_k^i-\Delta\theta_k^i
\bigr)
+
\alpha^i\widetilde R_k^i \\
&=
V_i^k\widetilde E_k^i
+
\alpha^i\widetilde R_k^i .
\end{aligned}
\]
This proves \autoref{eq:acc-error-rec}. Finally, unrolling the recursion with
\[
  \widetilde\Phi_{j,i}^k
  =
  V_j^kV_{j-1}^k\cdots V_i^k
\]
gives \autoref{eq:error-sum-acc}.
\end{proof}

\section{Finite-Horizon Error Bounds}
\label{app:error-bounds}
The bounds follow a common proof pattern: bound the local Taylor and
interaction residuals, control their propagation, and sum the resulting
contributions. We first prove the non-convex bounds used in the main
text, including possible amplification along the training trajectory.
The complementary strongly-convex analysis then shows how positive
curvature contracts errors. The final comparison explains how to
interpret both sets of bounds.\footnote{ChatGPT (OpenAI) assisted
with the analysis and exploration of ACC-SGD-IE error bounds in
this appendix.}
\subsection{Smooth Non-Convex Objectives}
\label{sec:noconvex-case}

Without strong convexity, propagation can amplify earlier residuals.
We first keep the trajectory-specific propagation factors explicit. We
then bound the true displacement using the total step budget and obtain
closed-form envelopes for both estimators.

\subsubsection{Assumptions and Propagation}
\label{app:nonconvex-propagation}

\begin{assumption}\label{asm:assumption-sgd-nonconvex}
Condition on the recorded batches and the coupled fixed-denominator
counterfactual. There exist finite $\Lambda,L,G>0$ such that, for each
$i<N$, every relevant sample $z\in Z^i$, and $\theta,\theta'$ on the
segment $\mathcal S_i^k=\{\theta^i+t\Delta\theta_k^i:0\le t\le1\}$,
\[
 \|H(z,\theta)\|_2\le\Lambda,
 \qquad
 \|H(z,\theta)-H(z,\theta')\|_2\le L\|\theta-\theta'\|_2.
\]
At target occurrences on the original trajectory,
$\|g(z_k,\theta^i)\|_2\le G$.
Each batch has size $M\ge1$ and contains the target at most once; use
nonnegative step sizes $\alpha^i\ge0$. Define
\[
\begin{aligned}
 \iota_i^k&=\mathbb{I}(z_k\in Z^i),\quad
 c_i^k=1-\frac{\iota_i^k}{M},\\
 d_i^k&=\|\Delta\theta_k^i\|_2,\quad
 A_i=\sum_{j=0}^{i-1}\alpha^j,
 \quad A_0=0.
\end{aligned}
\]
\end{assumption}

The accumulated budget $A_i$ tracks how much update time has elapsed.
It controls the displacement generated by repeated removal, including
possible amplification. The Hessian conditions apply along the full
segment between each observed iterate and its counterfactual iterate.

\begin{lemma}[Propagation Bound under Non-Convexity]
\label{lem:nonconvex-propagation}
Under \autoref{asm:assumption-sgd-nonconvex},
\[
 \|U_i\|_2\le1+\alpha^i\Lambda,
 \qquad
 \|V_i^k\|_2\le1+\alpha^i c_i^k\Lambda.
\]
Consequently,
\begin{align*}
 \|\Phi_{N-1,i+1}\|_2
 &\le\prod_{j=i+1}^{N-1}(1+\alpha^j\Lambda)
 \le e^{\Lambda(A_N-A_{i+1})},\\
 \|\widetilde\Phi_{N-1,i+1}^k\|_2
 &\le\prod_{j=i+1}^{N-1}(1+\alpha^j c_j^k\Lambda)
 \le e^{\Lambda(A_N-A_{i+1})}.
\end{align*}
\end{lemma}
\begin{proof}
The full averaged Hessian has norm at most $\Lambda$, whereas the
retained Hessian sum with denominator $M$ has norm at most
$c_i^k\Lambda$. Apply the triangle inequality to their update matrices,
then submultiplicativity and $1+t\le e^t$ for $t\ge0$.
The ordered products preserve the chronology of the updates.
\end{proof}

\subsubsection{Trajectory-Dependent Error Bounds}
\label{app:nonconvex-path-bounds}

These bounds retain the actual displacements, target occurrences and
propagation factors. They provide the starting point for the total-budget
envelopes derived below.

\begin{theorem}[Trajectory-dependent error bounds]
\label{thm:nonconvex-path-bounds}
Under \autoref{asm:assumption-sgd-nonconvex}, let
$p_i\ge\|U_i\|_2$ and $q_i^k\ge\|V_i^k\|_2$ be nonnegative
propagation bounds. Then
\begin{align}
 \|E_k^N\|_2
 &\le\sum_{i=0}^{N-1}\alpha^i
 \left(\prod_{j=i+1}^{N-1}p_j\right)
 \left(\frac{\iota_i^k\Lambda}{M}d_i^k+
 \frac{c_i^kL}{2}(d_i^k)^2\right),
 \label{eq:nonconvex-baseline-path}\\
 \|\widetilde E_k^N\|_2
 &\le\frac L2\sum_{i=0}^{N-1}\alpha^i c_i^k(d_i^k)^2
 \prod_{j=i+1}^{N-1}q_j^k.
 \label{eq:nonconvex-acc-path}
\end{align}
\end{theorem}
\begin{proof}
Taylor's theorem on $\mathcal S_i^k$ gives
$\|\widetilde R_k^i\|_2\le c_i^kL(d_i^k)^2/2$. The exact residual split
\[
 R_k^i=\widetilde R_k^i-
 \frac{\iota_i^k}{M}H(z_k,\theta^i)\Delta\theta_k^i
\]
therefore gives
$\|R_k^i\|_2\le\iota_i^k\Lambda d_i^k/M+c_i^kL(d_i^k)^2/2$.
Insert these into \autoref{eq:error-sum-classic} and
\autoref{eq:error-sum-acc}, and bound each ordered matrix product by the
product of its norms.
\end{proof}

\subsubsection{Controlling the True Displacement}
\label{app:nonconvex-displacement-control}

The following lemma bounds $d_i^k$ directly from the coupled updates.
Substituting this bound into the preceding theorem eliminates the
unknown counterfactual displacement from the final error envelopes.

\begin{lemma}[Displacement bound from the total step budget]
\label{lem:nonconvex-displacement}
Under \autoref{asm:assumption-sgd-nonconvex},
\begin{equation}\label{eq:nonconvex-displacement}
 d_i^k\le\frac{G}{M\Lambda}
 \left[\prod_{j=0}^{i-1}(1+\alpha^j\Lambda)-1\right]
 \le\frac{G}{M\Lambda}(e^{\Lambda A_i}-1).
\end{equation}
\end{lemma}
\begin{proof}
\textit{Step 1: Obtain a scalar displacement recurrence.}
Let
\[
 \overline H_i^k=\frac1M\sum_{z\in Z^i\setminus\{z_k\}}
 \int_0^1 H(z,\theta^i+t\Delta\theta_k^i)\,dt.
\]
The fundamental theorem of calculus gives the exact recursion
\[
 \Delta\theta_k^{i+1}
 =(I-\alpha^i\overline H_i^k)\Delta\theta_k^i+\Gamma_{k,i}.
\]
Since $\|\overline H_i^k\|_2\le c_i^k\Lambda$,
\[
\begin{aligned}
 d_{i+1}^k&\le(1+\alpha^i c_i^k\Lambda)d_i^k+
 \frac{\alpha^i\iota_i^kG}{M}\\
 &\le P_i d_i^k+\frac{\alpha^iG}{M},
 \qquad P_i=1+\alpha^i\Lambda.
\end{aligned}
\]
\textit{Step 2: Unroll and telescope.}
Unrolling from $d_0^k=0$ yields
\[
 d_i^k\le\frac GM\sum_{t=0}^{i-1}\alpha^t
 \prod_{j=t+1}^{i-1}P_j
 =\frac{G}{M\Lambda}\left(\prod_{j=0}^{i-1}P_j-1\right).
\]
The identity uses $P_t-1=\alpha^t\Lambda$ and telescopes suffix
products. Finally, $\prod_{j<i}P_j\le e^{\Lambda A_i}$.
This is the single-record case of a suppression-fraction bound with
$\iota_i^k/M\le r_*=1/M$.
\end{proof}

\subsubsection{Analysis for SGD-IE}
\label{app:sgd-nonconvex-case}

SGD-IE accumulates both the linear interaction residual and the
quadratic Taylor remainder. We bound these two contributions separately.

\begin{lemma}[First-Order Residual Bound for SGD-IE]
\label{lem:sgdie-nonconvex-residual}
Under \autoref{asm:assumption-sgd-nonconvex},
\[
 \|R_k^i\|_2\le\frac{\iota_i^k\Lambda}{M}d_i^k+
 \frac{c_i^kL}{2}(d_i^k)^2.
\]
\end{lemma}
\begin{proof}
This is the residual split and Taylor estimate used in the proof of
\autoref{thm:nonconvex-path-bounds}. Its linear term is first order in
the displacement, with the deletion coefficient fixed.
\end{proof}

\begin{theorem}[\sgdiehl{\parbox[t]{.64\columnwidth}{SGD-IE: First-Order Residual Accumulation under Non-Convexity}}]
\label{thm:sgd-nonconvex-bound}
Under \autoref{asm:assumption-sgd-nonconvex}, put $x=\Lambda A_N$.
The error for the removed training example $z_k$ satisfies
\begin{equation}\label{eq:error-nonconvex-correct}
\sgdiehl{
\begin{aligned}
 \|E_k^N\|_2&\le
 \frac{G}{M^2\Lambda}(xe^x-e^x+1)\\
 &\quad+\frac{LG^2}{2M^2\Lambda^3}(e^{2x}-1-2xe^x).
\end{aligned}
}
\end{equation}
\end{theorem}
\begin{proof}
\textit{Step 1: Bound the interaction contribution.}
Use \autoref{eq:nonconvex-baseline-path} with
$p_i=P_i=1+\alpha^i\Lambda$. Write $A=A_N$.
By \autoref{lem:nonconvex-displacement}, the interaction contribution is
at most
\[
 \frac{G}{M^2}e^{\Lambda A}
 \sum_{i=0}^{N-1}\alpha^i(1-e^{-\Lambda A_i}).
\]
Here we used $\iota_i^k\le1$ and the suffix bound
$\prod_{j>i}P_j\le e^{\Lambda(A-A_i)}$.
Since $t\mapsto1-e^{-\Lambda t}$ is nondecreasing, its left Riemann
sum is bounded by its integral over $[0,A]$, giving
\begin{equation}\label{eq:nonconvex-interaction-budget}
 \frac{G}{M^2}e^{\Lambda A}
 \int_0^A(1-e^{-\Lambda t})\,dt
 =\frac{G}{M^2\Lambda}(\Lambda A e^{\Lambda A}-e^{\Lambda A}+1).
\end{equation}
\textit{Step 2: Bound the Taylor contribution.}
Use $c_i^k\le1$ and the same suffix and
displacement bounds. It is at most
\[
 \frac{LG^2}{2M^2\Lambda^2}e^{\Lambda A}
 \sum_{i=0}^{N-1}\alpha^i
 (e^{\Lambda A_i}-2+e^{-\Lambda A_i}).
\]
The function $t\mapsto e^{\Lambda t}-2+e^{-\Lambda t}$ is also
nondecreasing for $t\ge0$. A second integral comparison yields
\begin{equation}\label{eq:nonconvex-taylor-budget}
\begin{aligned}
 &\frac{LG^2}{2M^2\Lambda^2}e^{\Lambda A}
 \int_0^A(e^{\Lambda t}-2+e^{-\Lambda t})\,dt\\
 &\qquad=\frac{LG^2}{2M^2\Lambda^3}
 (e^{2\Lambda A}-1-2\Lambda A e^{\Lambda A}).
\end{aligned}
\end{equation}
Adding \autoref{eq:nonconvex-interaction-budget} and
\autoref{eq:nonconvex-taylor-budget} proves the theorem.
\end{proof}

\subsubsection{Analysis for ACC-SGD-IE}
\label{app:acc-nonconvex-case}

ACC-SGD-IE retains the interaction in its update matrix, leaving only
the Taylor contribution in the residual. The same displacement bound
therefore yields the corresponding quadratic-residual envelope.

\begin{lemma}[Second-Order Residual Bound for ACC-SGD-IE]
\label{lem:residual-bound}
Under \autoref{asm:assumption-sgd-nonconvex},
\[
 \|\widetilde R_k^i\|_2\le\frac{c_i^kL}{2}(d_i^k)^2.
\]
\end{lemma}
\begin{proof}
Each of the $Mc_i^k$ retained gradient Taylor remainders has norm at
most $L(d_i^k)^2/2$; the original denominator $M$ gives the result.
\end{proof}

\begin{theorem}[\acchl{\parbox[t]{.64\columnwidth}{ACC-SGD-IE: Second-Order Residual Accumulation under Non-Convexity}}]
\label{thm:nonconvex-bound-acc-sgdie}
Under \autoref{asm:assumption-sgd-nonconvex}, put $x=\Lambda A_N$.
The error for the removed training example $z_k$ satisfies
\begin{equation}\label{eq:error-nonconvex-acc}
\acchl{
 \|\widetilde E_k^N\|_2\le
 \frac{LG^2}{2M^2\Lambda^3}(e^{2x}-1-2xe^x).
}
\end{equation}
\end{theorem}
\begin{proof}
In \autoref{eq:nonconvex-acc-path}, use
$q_i^k=1+\alpha^i c_i^k\Lambda\le P_i$. The Taylor contribution is
then bounded by the same displacement and integral calculation in
\autoref{eq:nonconvex-taylor-budget}. There is no additional interaction
residual in the corrected error recursion.
\end{proof}

\paragraph{Interpreting the total step budget.}
For small $A_N$, the interaction and Taylor envelopes have leading terms
$G\Lambda A_N^2/(2M^2)$ and $LG^2A_N^3/(6M^2)$, respectively. The
correction removes the former term from the common upper envelope.
\appref{subsec:compare-sgd-ie-new} explains how these upper bounds
relate to realised errors.

With the horizon-normalized schedule $\alpha^i=\gamma/\sqrt N$, the
total budget is $A_N=\gamma\sqrt N$. Thus the budget grows with the
horizon even though individual steps shrink. Controlling per-step
learning rates alone does not establish convergence of the finite-deletion
estimation error.

\subsection{Smooth Strongly Convex Objectives}
\label{sec:convex-case}

Strong convexity contracts propagated errors. We first bound the update
matrices and local residuals, then combine them using the exact error
unrollings in \appref{app:error-recursions}. An optional target-gradient
bound at the end gives a uniform scale for the true displacement.

\subsubsection{Comparison and Proof Guide}
\label{app:convex-comparison}
The following comparison collects the two error bounds in one place.
The first term in the SGD-IE bound comes from repeated deletion;
ACC-SGD-IE accounts for that interaction in its update and leaves the
Taylor remainder. We establish the curvature assumptions and then
prove each bound in the subsections below.

\begin{theorem}[Strongly-convex error comparison]
\label{thm:strong}
Suppose every relevant per-sample Hessian satisfies
$\lambda I\preceq H(z,\theta)\preceq\Lambda I$ and is $L$-Lipschitz on
the corresponding original-to-counterfactual segment, as specified in
\autoref{asm:assumption-sgd-convex}. Let $M\ge2$ and use a constant step
size $0<\alpha\le1/\Lambda$. Then
\[
\sgdiehl{
 \|E_k^N\|_2\le\sum_{i=0}^{N-1}
 \alpha(1-\alpha\lambda)^{N-1-i}
 \left(\frac{\iota_i^k\Lambda}{M}d_i^k+
 \frac{c_i^kL}{2}(d_i^k)^2\right),
}
\]
and
\[
\acchl{
 \|\widetilde E_k^N\|_2\le\sum_{i=0}^{N-1}
 \frac{\alpha c_i^k L}{2}(d_i^k)^2
 \prod_{j=i+1}^{N-1}(1-\alpha\lambda c_j^k).
}
\]
If $d_i^k\le D_k$ for all $i<N$, then
\[
\sgdiehl{
\begin{aligned}
 \|E_k^N\|_2&\le\frac{\Lambda D_k}{M}
 \sum_{i=0}^{N-1}\alpha(1-\alpha\lambda)^{N-1-i}\iota_i^k\\
 &\quad+\frac{LD_k^2}{2\lambda}[1-(1-\alpha\lambda)^N],
\end{aligned}
}
\]
whereas
\[
\acchl{
 \|\widetilde E_k^N\|_2\le\frac{LD_k^2}{2\lambda}
 \left[1-\prod_{i=0}^{N-1}(1-\alpha\lambda c_i^k)\right]
 \le\frac{LD_k^2}{2\lambda}.
}
\]
\end{theorem}

\noindent\textbf{Interpretation.}
ACC-SGD-IE removes the interaction residual linear in $d_i^k$, leaving
a Taylor residual quadratic in $d_i^k$ at a fixed deletion coefficient
$1/M$. The linear term arises when the target reappears after an earlier
removal has created a nonzero displacement. Including this term in
$V_i^k$ accounts for the effect of the new removal on the accumulated
state. The retained fraction $c_i^k$ both reduces the Taylor remainder
and weakens contraction. Accounting for both effects yields the uniform
ACC-SGD-IE bound above.

\subsubsection{Assumptions and Shared Bounds}
\label{app:convex-shared-bounds}

\begin{assumption}\label{asm:assumption-sgd-convex}
Fix a removed training example $z_k$ and the recorded SGD batches. For
$i=0,\ldots,N-1$, let
\[
 \mathcal S_i^k=\{\theta^i+t\Delta\theta_k^i:0\le t\le1\}
\]
be the original-to-counterfactual displacement segment. There exist
$\lambda,\Lambda,L>0$ such that, for every $z\in Z^i$ and
$\theta,\theta'\in\mathcal S_i^k$,
\[
 \lambda I\preceq H(z,\theta)\preceq\Lambda I,
 \qquad
 \|H(z,\theta)-H(z,\theta')\|_2\le L\|\theta-\theta'\|_2.
\]
Each mini-batch has size $M\ge2$ and contains the target at most once.
Use $\alpha^i\equiv\alpha$ with $0<\alpha\le1/\Lambda$, and write
\[
 \iota_i^k=\mathbb{I}(z_k\in Z^i),\qquad
 c_i^k=1-\frac{\iota_i^k}{M},\qquad
 d_i^k=\|\Delta\theta_k^i\|_2.
\]
\end{assumption}

These assumptions concern the curvature of each relevant sample along
the segment connecting the observed and counterfactual iterates. The
following bounds use the true displacement $d_i^k$; the additional
gradient condition in \autoref{cor:strong-displacement-bound} provides a
uniform displacement bound when needed.

The first lemma controls propagation through the full mini-batch update.

\begin{lemma}[Linear contraction of $U_i$]\label{lem:U-contraction-shared}
Under \autoref{asm:assumption-sgd-convex},
\[
 \|U_i\|_2\le1-\alpha\lambda.
\]
\end{lemma}
\begin{proof}
The mini-batch Hessian at $\theta^i$ has eigenvalues in
$[\lambda,\Lambda]$. Since $\alpha\le1/\Lambda$, the eigenvalues of
$U_i=I-\alpha H(Z^i,\theta^i)$ are nonnegative and at most
$1-\alpha\lambda$.
\end{proof}

The next lemma separates the interaction residual, which is linear in
displacement, from the retained-sample Taylor remainder, which is quadratic.

\begin{lemma}[Residual bounds]\label{lem:residual-bounds-strong}
Under \autoref{asm:assumption-sgd-convex},
\begin{align}
 \|R_k^i\|_2
 &\le\frac{\iota_i^k\Lambda}{M}d_i^k+
 \frac{c_i^kL}{2}(d_i^k)^2,
 \label{eq:sgdie-residual-bound-strong}\\
 \|\widetilde R_k^i\|_2
 &\le\frac{c_i^kL}{2}(d_i^k)^2.
 \label{eq:acc-residual-bound-strong}
\end{align}
\end{lemma}
\begin{proof}
For every retained sample, the integral form of Taylor's theorem gives
\[
\begin{gathered}
 g(z,\theta_k^i)-g(z,\theta^i)
 =H(z,\theta^i)\Delta\theta_k^i+r_z^i,\\
 r_z^i=\int_0^1
 [H(z,\theta^i+t\Delta\theta_k^i)-H(z,\theta^i)]
 \Delta\theta_k^i\,dt.
\end{gathered}
\]
Hence $\|r_z^i\|_2\le L(d_i^k)^2/2$. There are $Mc_i^k$ retained
samples, so
\[
 \widetilde R_k^i=\frac1M\sum_{z\in Z^i\setminus\{z_k\}}r_z^i,
 \qquad
 R_k^i=\widetilde R_k^i-
 \frac{\iota_i^k}{M}H(z_k,\theta^i)\Delta\theta_k^i.
\]
The triangle inequality and $\|H(z_k,\theta^i)\|_2\le\Lambda$ prove
both bounds. Keeping the retained-sample fraction $c_i^k$ also allows
the ACC-SGD-IE uniform bound to telescope exactly.
\end{proof}

ACC-SGD-IE propagates through the retained-batch update. Its contraction
rate includes the same retained-sample fraction $c_i^k$.

\begin{lemma}[Linear contraction of $V_i^k$]\label{lem:V-contraction-strong}
Under \autoref{asm:assumption-sgd-convex},
\[
 \|V_i^k\|_2\le1-\alpha\lambda c_i^k.
\]
\end{lemma}
\begin{proof}
Since the original denominator is retained,
\[
 V_i^k=I-\frac\alpha M\sum_{z\in Z^i\setminus\{z_k\}}H(z,\theta^i).
\]
The retained Hessian sum divided by $M$ has eigenvalues in
$[c_i^k\lambda,c_i^k\Lambda]$. The step-size condition makes the update
matrix positive semidefinite, giving the stated norm bound. This also
covers $\iota_i^k=0$, when $c_i^k=1$ and $V_i^k=U_i$.
\end{proof}

\subsubsection{Analysis for SGD-IE}
\label{app:error-analysis-appendix-SGD-IE}

Combining full-batch contraction with the two-part residual bound gives
the trajectory-dependent bound and its uniform-displacement form.

\begin{theorem}[\sgdiehl{SGD-IE: first-order residual accumulation}]
\label{thm:sgdie-strong-corrected}
Under \autoref{asm:assumption-sgd-convex},
\begin{equation}\label{eq:sgdie-strong-corrected}
\sgdiehl{
 \|E_k^N\|_2\le\sum_{i=0}^{N-1}
 \alpha(1-\alpha\lambda)^{N-1-i}
 \left(\frac{\iota_i^k\Lambda}{M}d_i^k+
 \frac{c_i^kL}{2}(d_i^k)^2\right).
}
\end{equation}
If $d_i^k\le D_k$ for all $i<N$, then
\begin{equation}\label{eq:sgdie-strong-uniform}
\sgdiehl{
\begin{aligned}
 \|E_k^N\|_2&\le\frac{\Lambda D_k}{M}
 \sum_{i=0}^{N-1}\alpha(1-\alpha\lambda)^{N-1-i}\iota_i^k\\
 &\quad+\frac{LD_k^2}{2\lambda}[1-(1-\alpha\lambda)^N].
\end{aligned}
}
\end{equation}
\end{theorem}
\begin{proof}
\textit{Step 1: Bound the propagated residuals.}
Unroll \autoref{eq:error-recursion-classic} from $E_k^0=0$.
Submultiplicativity and \autoref{lem:U-contraction-shared} give
\[
 \|\Phi_{N-1,i+1}\|_2\le(1-\alpha\lambda)^{N-1-i}.
\]
Combining this with \autoref{eq:sgdie-residual-bound-strong} proves
\autoref{eq:sgdie-strong-corrected}.

\textit{Step 2: Sum the contraction weights.}
For the uniform form, use
$c_i^k\le1$, $d_i^k\le D_k$, and
\[
 \sum_{i=0}^{N-1}\alpha(1-\alpha\lambda)^{N-1-i}
 =\frac{1-(1-\alpha\lambda)^N}{\lambda}.
\]
\end{proof}

\subsubsection{Analysis for ACC-SGD-IE}
\label{subsec:acc-convex-case}

The corrected recurrence retains only the quadratic Taylor residual.
Its retained-batch contraction yields a uniform bound by telescoping
suffix products.

\begin{theorem}[\acchl{\parbox[t]{.64\columnwidth}{ACC-SGD-IE: Second-Order Residual Accumulation}}]
\label{thm:acc-sgdie-strong-corrected}
Under \autoref{asm:assumption-sgd-convex},
\begin{equation}\label{eq:acc-bound-strong-corrected}
\acchl{
 \|\widetilde E_k^N\|_2\le\sum_{i=0}^{N-1}
 \frac{\alpha c_i^kL}{2}(d_i^k)^2
 \prod_{j=i+1}^{N-1}(1-\alpha\lambda c_j^k).
}
\end{equation}
If $d_i^k\le D_k$ for all $i<N$, then
\begin{equation}\label{eq:acc-bound-strong-uniform}
\acchl{
 \|\widetilde E_k^N\|_2\le\frac{LD_k^2}{2\lambda}
 \left[1-\prod_{i=0}^{N-1}(1-\alpha\lambda c_i^k)\right]
 \le\frac{LD_k^2}{2\lambda}.
}
\end{equation}
\end{theorem}
\begin{proof}
\textit{Step 1: Bound the propagated Taylor residuals.}
The exact error unrolling in \autoref{eq:error-sum-acc}, together with
\autoref{lem:V-contraction-strong} and
\autoref{eq:acc-residual-bound-strong}, proves
\autoref{eq:acc-bound-strong-corrected}.

\textit{Step 2: Telescope the retained-batch contraction weights.}
For the uniform form, put
$q_i^k=1-\alpha\lambda c_i^k$. Then
\[
 \sum_{i=0}^{N-1}\alpha c_i^k\prod_{j=i+1}^{N-1}q_j^k
 =\frac1\lambda\sum_{i=0}^{N-1}(1-q_i^k)
 \prod_{j=i+1}^{N-1}q_j^k
 =\frac{1-\prod_{i=0}^{N-1}q_i^k}{\lambda}.
\]
The last equality telescopes consecutive suffix products. Inserting
$d_i^k\le D_k$ proves \autoref{eq:acc-bound-strong-uniform}.
\end{proof}

\subsubsection{A Uniform Bound on the True Displacement}
\label{app:convex-displacement}

A bound on the target gradient at its observed occurrences controls
the displacement scale $D_k$ used above.

\begin{corollary}[A local bound on the true displacement]
\label{cor:strong-displacement-bound}
In addition to \autoref{asm:assumption-sgd-convex}, suppose
$\|g(z_k,\theta^i)\|_2\le G$ at target occurrences on the original
trajectory. Then
\[
 d_i^k\le\frac{G}{(M-1)\lambda},\qquad i=0,\ldots,N.
\]
\end{corollary}
\begin{proof}
Define the segment-averaged retained Hessian
\[
 \overline H_i^k=\frac1M\sum_{z\in Z^i\setminus\{z_k\}}
 \int_0^1 H(z,\theta^i+t\Delta\theta_k^i)\,dt.
\]
Its eigenvalues lie in $[c_i^k\lambda,c_i^k\Lambda]$. The exact
gradient-difference identity gives
\[
\begin{aligned}
 \Delta\theta_k^{i+1}
 &=(I-\alpha\overline H_i^k)\Delta\theta_k^i+\Gamma_{k,i},\\
 d_{i+1}^k&\le(1-\alpha c_i^k\lambda)d_i^k
 +\frac{\alpha\iota_i^kG}{M}.
\end{aligned}
\]
As $c_i^k\ge1-1/M$ and $\iota_i^k\le1$, induction from $d_0^k=0$
with $D_k=G/[(M-1)\lambda]$ proves the claim. The gradient bound is
needed only for the target at its occurrences on the observed trajectory.
\end{proof}

\subsection{Comparing the Error Envelopes}
\label{subsec:compare-sgd-ie-new}

Both methods approximate the same coupled, fixed-denominator
finite-deletion displacement. SGD-IE gives the sample-weight derivative
at zero removal weight (\appref{app:weight-tangent-proof}); ACC-SGD-IE
uses the retained-batch Jacobian to propagate the accumulated finite
displacement. The exact residual split makes the effect of this
correction explicit:
\[
 R_k^i=\widetilde R_k^i-
 \frac{\iota_i^k}{M}H(z_k,\theta^i)\Delta\theta_k^i.
\]
The correction absorbs the interaction term into $V_i^k$, leaving the
retained-data Taylor remainder. We compare the resulting envelopes first
under contractive strong-convexity assumptions, then under non-convex
propagation bounds. Here, residual order refers to powers of the state
displacement at a fixed finite-deletion coefficient; its distinction
from deletion-weight Taylor order is explained in
\appref{app:weight-tangent-proof}.

\subsubsection{Smooth strongly--convex objectives}

Under \autoref{asm:assumption-sgd-convex}, the per-sample Hessian bounds
hold on every relevant true displacement segment. Write
$d_i^k=\|\Delta\theta_k^i\|_2$ and $c_i^k=1-\iota_i^k/M$.
\autoref{thm:sgdie-strong-corrected} and
\autoref{thm:acc-sgdie-strong-corrected} give
\begin{align}
 \|E_k^N\|_2
 &\le\sum_{i=0}^{N-1}\alpha(1-\alpha\lambda)^{N-1-i}
 \left(\frac{\iota_i^k\Lambda}{M}d_i^k+
 \frac{c_i^kL}{2}(d_i^k)^2\right),
 \label{eq:sgdie-strong-compare}\\
 \|\widetilde E_k^N\|_2
 &\le\sum_{i=0}^{N-1}\frac{\alpha c_i^kL}{2}(d_i^k)^2
 \prod_{j=i+1}^{N-1}(1-\alpha\lambda c_j^k).
 \label{eq:accsgdie-strong-compare}
\end{align}

With $1/M$ fixed, the interaction residual is linear in the true
displacement, whereas the Taylor remainder is quadratic. A first
occurrence starting from zero displacement contributes no interaction
residual. Later occurrences can interact with displacement accumulated
from earlier deletion steps.

To compare upper envelopes using the same propagation factor, put
$q_i^k=1-\alpha\lambda c_i^k\ge1-\alpha\lambda$. Both errors are then
bounded by the respective quantities
\begin{align*}
 \mathcal B_{\mathrm{ACC}}
 &=\sum_{i=0}^{N-1}\frac{\alpha c_i^kL}{2}(d_i^k)^2
 \prod_{j=i+1}^{N-1}q_j^k,\\
 \mathcal B_{\mathrm{SGD}}
 &=\mathcal B_{\mathrm{ACC}}+
 \sum_{i=0}^{N-1}\frac{\alpha\iota_i^k\Lambda}{M}d_i^k
 \prod_{j=i+1}^{N-1}q_j^k.
\end{align*}
The common envelopes satisfy
$\mathcal B_{\mathrm{ACC}}\le\mathcal B_{\mathrm{SGD}}$: ACC-SGD-IE removes
the additional interaction contribution. This ordering concerns upper
bounds.

If $d_i^k\le D_k$ for all $i<N$, the uniform bounds are
\[
 \begin{aligned}
 \|E_k^N\|_2&\le\frac{\Lambda D_k}{M}
 \sum_{i=0}^{N-1}\alpha(1-\alpha\lambda)^{N-1-i}\iota_i^k\\
 &\quad+\frac{LD_k^2}{2\lambda}[1-(1-\alpha\lambda)^N],
 \end{aligned}
\]
and
\[
 \|\widetilde E_k^N\|_2\le\frac{LD_k^2}{2\lambda}
 \left[1-\prod_{i=0}^{N-1}(1-\alpha\lambda c_i^k)\right].
\]
The retained-sample fraction $c_i^k$ makes the Taylor contribution
telescope against retained-batch contraction, avoiding the looser
$1/(1-1/M)$ factor. If target gradients at
observed iterates are bounded by $G$,
\autoref{cor:strong-displacement-bound} supplies
$D_k=G/[(M-1)\lambda]$. With other constants fixed, both uniform
envelopes scale as $O(M^{-2})$ with batch size. The correction therefore
changes the residual sources within this common batch-size order.

\subsubsection{Smooth non-convex objectives}

Under \autoref{asm:assumption-sgd-nonconvex}, define
$A_i=\sum_{j<i}\alpha^j$ and $x=\Lambda A_N$.
\autoref{thm:sgd-nonconvex-bound} and
\autoref{thm:nonconvex-bound-acc-sgdie} give
\begin{align}
 \|E_k^N\|_2
 &\le\frac{G}{M^2\Lambda}(xe^x-e^x+1)
 +\frac{LG^2}{2M^2\Lambda^3}(e^{2x}-1-2xe^x),
 \label{eq:sgdie-nonconvex-compare}\\
 \|\widetilde E_k^N\|_2
 &\le\frac{LG^2}{2M^2\Lambda^3}(e^{2x}-1-2xe^x).
 \label{eq:accsgdie-nonconvex-compare}
\end{align}
These are worst-case total-budget envelopes for the single-record
suppression fraction $\iota_i^k/M\le1/M$. The sharper path bounds in
\autoref{thm:nonconvex-path-bounds} retain the true occurrence pattern,
displacements, and separate propagation products.

For small $A_N$, the interaction envelope has leading term
$G\Lambda A_N^2/(2M^2)$, while the common Taylor envelope has leading
term $LG^2 A_N^3/(6M^2)$. The correction removes the lower-order
interaction term from this common upper envelope. This identifies a
regime in which ACC-SGD-IE can improve the finite-deletion approximation:
the interaction is substantial relative to the retained-data Taylor
remainder.

The derived displacement enclosure
\[
 d_i^k\le\frac{G}{M\Lambda}(e^{\Lambda A_i}-1)
\]
accounts for repeated deletion and possible non-convex amplification.
With $\alpha^i=\gamma/\sqrt N$, the total budget grows as
$A_N=\gamma\sqrt N$, so these envelopes can increase with the horizon.
Their dependence on the accumulated budget explains why longer training
must be assessed together with the occurrence pattern, propagation and
remaining Taylor residual. The trajectory-dependent bounds in
\autoref{thm:nonconvex-path-bounds} retain these quantities explicitly.

\section{Implementation and Computational Cost}
\label{sec:acc_implementation}

\subsection{Full Displacements and Shared Scalar Queries}
\label{app:complexity}

We measure propagation cost in mini-batch Hessian--vector products (HVPs).
An HVP applies curvature to a vector without forming the Hessian. Let \(N\) denote the number of SGD updates, \(n\) the number of
training examples, \(M\) the mini-batch size, and \(p\) the number of model
parameters.

\subsubsection{Classical SGD--IE}
For a target example \(z_k\), classical SGD--IE propagates one influence vector
along the recorded training trajectory. At each step, applying the local
linearized update operator \(U_i\) requires one mini-batch HVP without explicitly
forming the Hessian. The per-target cost is therefore \(\Theta(N)\) HVPs. A
naive all-sample implementation that repeats this procedure for all training
examples costs \(\Theta(Nn)\) HVPs.
For a common scalar query $v^\top\widehat{\Delta\theta_k^N}$ with fixed
$v$, classical SGD--IE can instead share a backward adjoint pass using
$O(N)$ mini-batch HVPs, followed by gradient readouts at the target
occurrences. These costs correspond to two outputs: $\Theta(Nn)$ HVPs for all
full-parameter displacement vectors, and a shared $O(N)$-HVP pass for
a common scalar query, followed by occurrence-wise gradient readouts.

\subsubsection{ACC--SGD--IE}
ACC--SGD--IE maintains a propagated influence state \(\widetilde{\Delta\theta}_k^i\) for each target
example. For a single queried example, the state is transported along the same
trajectory and receives an occurrence-level correction whenever \(z_k\in Z^i\).
Thus, the per-target complexity remains \(\Theta(N)\) HVPs up to a constant
factor determined by the number of re-occurrences.

For all-sample attribution in the full parameter space, ACC--SGD--IE propagates
\(n\) distinct states. At step \(i\), the mini-batch Hessian is shared but acts
on different vectors \(\{\widetilde{\Delta\theta}_k^i\}_{k=1}^n\). Hence, a direct implementation has
per-step work
\[
(n-M)\times \mathrm{HVP}\bigl(H(Z^i,\theta^i),\widetilde{\Delta\theta}_k^i\bigr)
\;+\;
M\times \mathrm{HVP}\bigl(H_{\mathrm{ret},i}^k,\widetilde{\Delta\theta}_k^i\bigr),
\]
where $H_{\mathrm{ret},i}^k=M^{-1}\sum_{z\in Z^i\setminus\{z_k\}}H(z,\theta^i)$ and the second term uses the retained-batch Hessian for examples in the current mini-batch. Equivalently, applying the full-batch operator and its explicit correction requires $n$ batch HVPs and at most $M$ sample HVPs per step. The total cost is therefore
\[
\Theta(Nn)
\]
vector-HVP applications. Storing all full-parameter states simultaneously
requires \(O(np)\) memory.

In practice, we process target examples in shards of size \(K\), reducing the
target-state memory requirement to \(O(Kp)\), in addition to any storage
used for the recorded training trajectory. The total number of vector-HVP applications
across all shards remains \(\Theta(Nn)\). Batched HVPs implemented with
\texttt{torch.func.vmap} improve wall-clock efficiency by fusing multiple
vector-HVP applications, while preserving the same asymptotic scaling.
The ACC--SGD--IE transport $V_i^k$ depends on the target, so the classical
shared adjoint above does not directly provide all ACC scalar scores.

\subsection{Implementation Choices and Approximation Scope}
\label{app:acceleration}

We process targets in shards and vectorize HVPs with
\texttt{torch.func.vmap}. The implementation also supports restricting
propagation to selected layers, such as the final fully connected layer;
this estimates influence in the chosen representation. Sharding and
batching preserve the propagated vectors, whereas layer restriction,
projection, trajectory segmentation, and proxy models alter the
approximation.

\section{Additional Experiments and Evaluation Protocols}
\label{app:additional-experiments}
The additional experiments examine two aspects of accumulative correction:
accuracy with convex models, behavior across training configurations. We first explain the
shared evaluation design, then report the extended results.

\subsection{Evaluation Tasks and Ground Truth}
\label{sec:rational}
We follow the SGD-IE evaluation protocol~\citep{hara2019data}, using models for which leave-one-out retraining supplies ground-truth effects. This design allows direct comparison of estimation accuracy and ranking fidelity under a controlled training schedule. The main experiments assess influence estimation, multi-epoch behavior, data cleansing, and industrial finite-deletion accuracy. The appendix provides complementary convex-model results and extended cross-epoch comparisons.

\subsection{Loss-Change Estimation with Convex Models} \label{app:results_convex}
\begin{figure*}[!t]
  \centering
  \begin{minipage}[t]{0.49\textwidth}
    \centering
    \includegraphics[width=\linewidth]{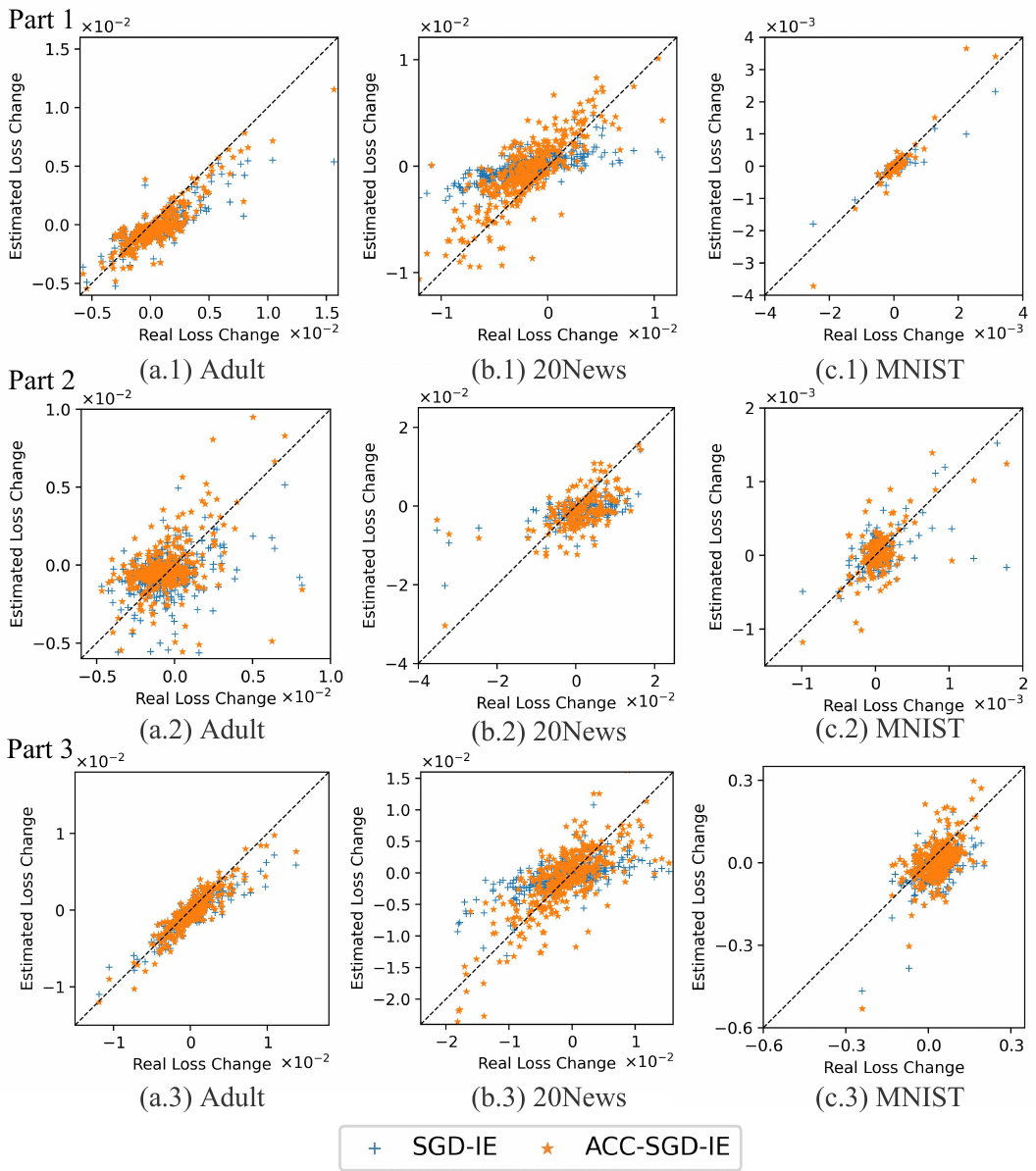}\par
    {\small (a) Non-convex regime.}
  \end{minipage}\hfill
  \begin{minipage}[t]{0.49\textwidth}
    \centering
    \includegraphics[width=\linewidth]{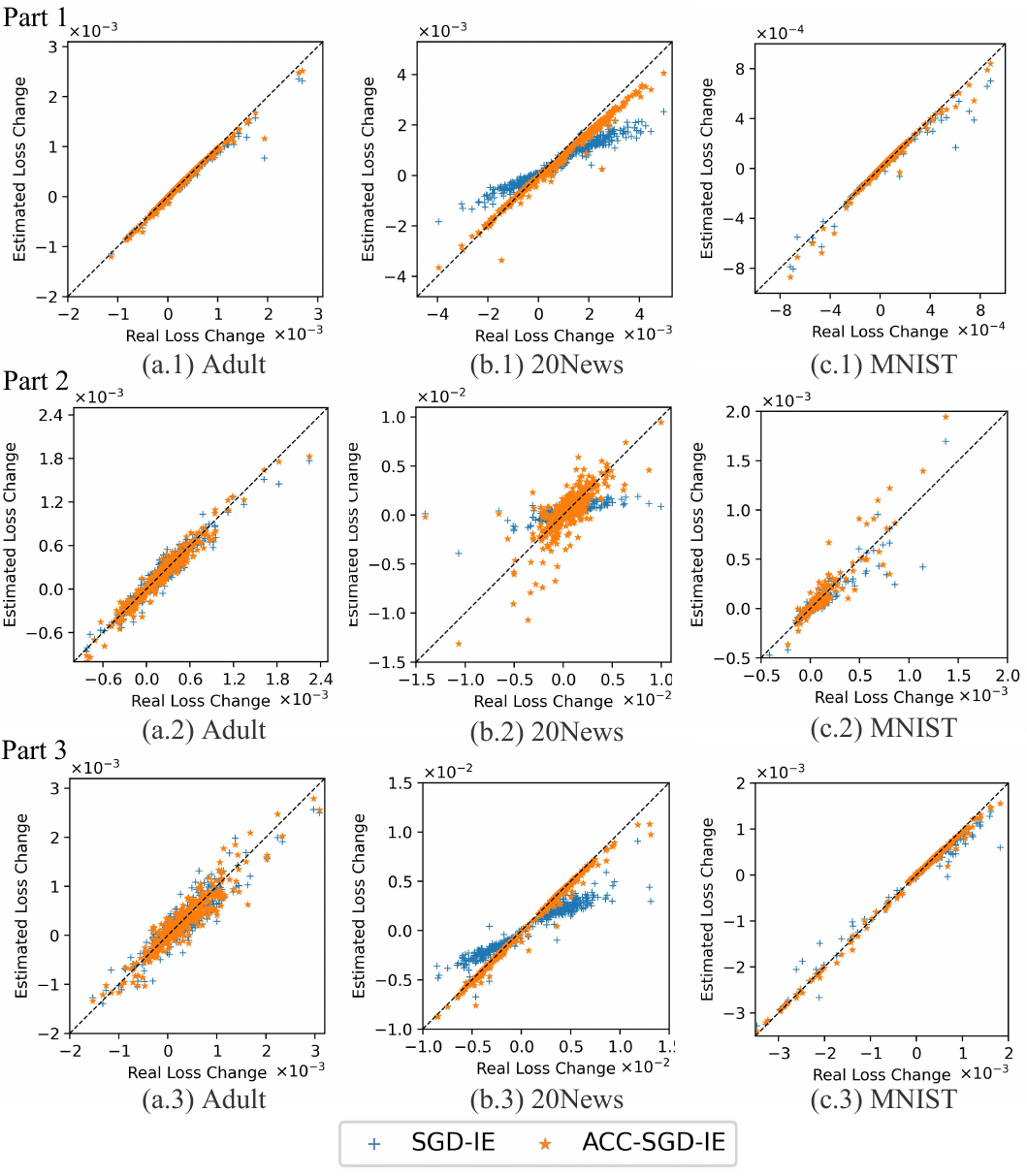}\par
    {\small (b) Convex regime.}
  \end{minipage}
  \caption{Loss change estimation under non-convex (left) and convex (right) regimes. Results on clean, feature-noisy, and label-noisy data. Points closer to the black diagonal indicate more accurate estimates of the real loss change. Parts 1--3 correspond respectively to the entries in \autoref{tab:all_influence} and \autoref{tab:all_influence_convex}.}
  \label{fig:estimation_plot}

  \par\medskip
  \centering
  \includegraphics[width=0.9\textwidth]{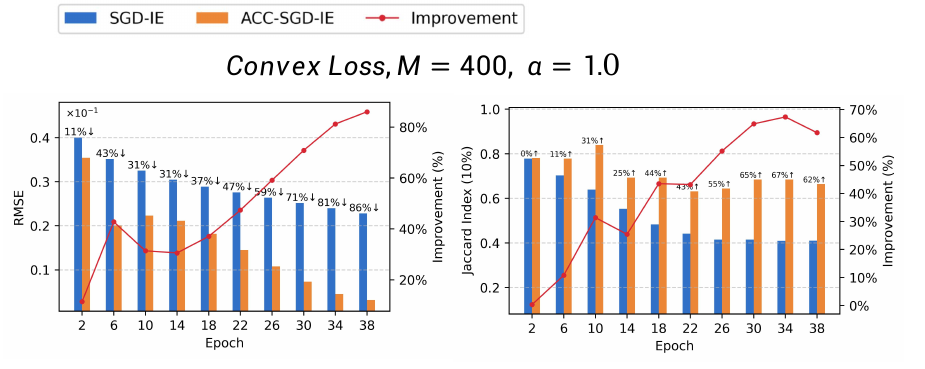}
  \caption{Highlighted cross-epoch convex controls on 20News with
  logistic regression. Left: loss-change RMSE. Right: Jaccard@10\%.
  Blue bars denote SGD-IE, orange bars ACC-SGD-IE, and red curves the
  relative improvement.}
  \label{fig:convex-epoch-highlight}
\end{figure*}

We use the datasets and feature/label-noise settings from
\autoref{subsec:exp_estimation}, with logistic regression as the convex
model. All datasets use learning rate $\gamma=0.1$, batch size $M=100$,
and 30 training epochs. Both estimators track influence from the first
epoch through the final epoch. We repeat each experiment 20 times;
\autoref{tab:all_influence_convex} reports means and standard deviations,
and \autoref{fig:estimation_plot} shows the corresponding scatter plots
for clean data, feature noise, and label noise. ACC--SGD--IE improves
estimation and ranking across nearly all reported convex settings,
complementing the non-convex results in the main text.

\begin{table*}[!t]
  \centering
  \caption{Comprehensive comparison of influence–estimation accuracy and robustness under strong convex regime. The best value for each metric within a setting is \textbf{bold}. We add i.i.d.\ noise $x \!\leftarrow\! x+\mathcal{N}(0,\sigma^{2})$ to every feature on Adult \& MNIST. We randomly set a fraction $\sigma\times100\%$ of the zero entries in each one-hot word vector to 1 on 20News. A fraction $\rho\times 100\%$ of the training labels is flipped uniformly at random.}
  \label{tab:all_influence_convex}
  \resizebox{\linewidth}{!}{\sisetup{detect-all}
  \begin{tabular}{@{}l c l l l l l l l@{}}
    \specialrule{1pt}{0pt}{0pt}
    \textbf{Dataset} & \textbf{Setting} & \textbf{Method} &
    {\textbf{RMSE}$\;\downarrow$} &
    {\textbf{Kendall's Tau}$\;\uparrow$} &
    {\textbf{Jacc@70\%}$\;\uparrow$} &
    {\textbf{Jacc@50\%}$\;\uparrow$} &
    {\textbf{Jacc@30\%}$\;\uparrow$} &
    {\textbf{Jacc@10\%}$\;\uparrow$} \\
    \midrule[0.8pt]
    \multicolumn{9}{c}{\textbf{Part 1.\; Estimation on Clean Data.}}\\
    \midrule[0.8pt]
    \multirow{2}{*}{Adult}
    & \multirow{2}{*}{-}
& SGD-IE & $1.00(1.00)\times10^{-4}$ & 0.9724(0.00) & 0.9616(0.01) & 0.9561(0.01) & 0.9306(0.02) & \textbf{0.9446(0.04)} \\
& & ACC-SGD-IE & \multicolumn{1}{l}{$\bm {1.00(1.00)\times10^{-4}}$} & \textbf{0.9774(0.00)} & \textbf{0.9627(0.01)} & \textbf{0.9608(0.01)} & \textbf{0.9410(0.02)} & 0.9369(0.05) \\

    \midrule
    \multirow{2}{*}{20News}
    & \multirow{2}{*}{-}
& SGD-IE & $1.04(1.25)\times10^{-2}$ & 0.9086(0.07) & 0.9038(0.08) & 0.8574(0.11) & 0.7832(0.18) & 0.6863(0.26) \\
& & ACC-SGD-IE & \multicolumn{1}{l}{$\bm {9.90(12.30)\times10^{-3}}$} & \textbf{0.9691(0.00)} & \textbf{0.9560(0.02)} & \textbf{0.9466(0.01)} & \textbf{0.9279(0.01)} & \textbf{0.9357(0.02)} \\

    \midrule
    \multirow{2}{*}{MNIST}
    & \multirow{2}{*}{-}
& SGD-IE & $2.50(2.80)\times10^{-3}$ & 0.9739(0.00) & 0.9806(0.00) & 0.9465(0.01) & 0.9395(0.01) & 0.9063(0.05) \\
& & ACC-SGD-IE & \multicolumn{1}{l}{$\bm {2.40(2.80)\times10^{-3}}$} & \textbf{0.9851(0.00)} & \textbf{0.9876(0.00)} & \textbf{0.9754(0.01)} & \textbf{0.9633(0.01)} & \textbf{0.9634(0.02)} \\
    \midrule
\multicolumn{3}{c}{\textbf{Average Improve (\%)}} 
  & \multicolumn{1}{c}{$\downarrow2.94\%$}
  & \multicolumn{1}{c}{$\uparrow2.78\%$}
  & \multicolumn{1}{c}{$\uparrow2.20\%$}
  & \multicolumn{1}{c}{$\uparrow4.65\%$}
  & \multicolumn{1}{c}{$\uparrow7.38\%$}
  & \multicolumn{1}{c}{$\uparrow13.94\%$} \\
    \midrule[0.8pt]
    \multicolumn{9}{c}{\textbf{Part 2.\; Estimation on Data Corrupted by Feature Noise.}}\\
    \midrule[0.8pt]
    \multirow{4}{*}{Adult}
        & \multirow{2}{*}{$\sigma=0.01$}
& SGD-IE & $3.00(2.00)\times10^{-4}$ & \textbf{0.6863(0.28)} & 0.7565(0.14) & \textbf{0.7001(0.21)} & \textbf{0.6385(0.23)} & 0.5776(0.23) \\
& & ACC-SGD-IE & \multicolumn{1}{l}{$\bm {3.00(3.00)\times10^{-4}}$} & 0.6782(0.29) & \textbf{0.7580(0.13)} & 0.6940(0.21) & 0.6274(0.24) & \textbf{0.5777(0.28)} \\

        \cmidrule{2-9}
        & \multirow{2}{*}{$\sigma=0.05$}
& SGD-IE & $2.00(0.00)\times10^{-4}$ & 0.7343(0.08) & 0.7205(0.06) & 0.6722(0.08) & 0.6264(0.08) & 0.5677(0.12) \\
& & ACC-SGD-IE & \multicolumn{1}{l}{$\bm {2.00(0.00)\times10^{-4}}$} & \textbf{0.7353(0.08)} & \textbf{0.7226(0.05)} & \textbf{0.6731(0.06)} & \textbf{0.6423(0.08)} & \textbf{0.5722(0.11)} \\
    \midrule
    \multirow{2}{*}{20News}
        & \multirow{2}{*}{$\sigma=2.5\times10^{-3}$}
& SGD-IE & $9.00(10.00)\times10^{-4}$ & 0.1754(0.04) & 0.5571(0.01) & 0.3751(0.02) & 0.2247(0.01) & 0.0991(0.02) \\
& & ACC-SGD-IE & \multicolumn{1}{l}{$\bm {7.00(6.00)\times10^{-4}}$} & \textbf{0.1859(0.04)} & \textbf{0.5615(0.01)} & \textbf{0.3828(0.01)} & \textbf{0.2422(0.03)} & \textbf{0.1064(0.02)} \\
    \midrule
    \multirow{4}{*}{MNIST}
        & \multirow{2}{*}{$\sigma=0.01$}
& SGD-IE & $2.00(1.00)\times10^{-4}$ & \textbf{0.3834(0.09)} & 0.6353(0.01) & 0.5117(0.02) & \textbf{0.4432(0.02)} & 0.3370(0.06) \\
& & ACC-SGD-IE & \multicolumn{1}{l}{$\bm {2.00(1.00)\times10^{-4}}$} & 0.3748(0.10) & \textbf{0.6469(0.03)} & \textbf{0.5285(0.03)} & 0.4154(0.03) & \textbf{0.3617(0.09)} \\
        \cmidrule{2-9}
        & \multirow{2}{*}{$\sigma=0.05$} 
& SGD-IE & $1.00(0.00)\times10^{-4}$ & 0.6803(0.06) & 0.6904(0.04) & 0.6345(0.05) & 0.5882(0.08) & 0.5417(0.07) \\
& & ACC-SGD-IE & \multicolumn{1}{l}{$\bm {1.00(0.00)\times10^{-4}}$} & \textbf{0.6961(0.06)} & \textbf{0.7061(0.03)} & \textbf{0.6523(0.05)} & \textbf{0.5932(0.09)} & \textbf{0.5766(0.06)} \\
    \midrule
\multicolumn{3}{c}{\textbf{Average Improve (\%)}}

  & \multicolumn{1}{c}{$\downarrow4.44\%$}
  & \multicolumn{1}{c}{$\uparrow1.00\%$}
  & \multicolumn{1}{c}{$\uparrow1.08\%$}
  & \multicolumn{1}{c}{$\uparrow1.48\%$}
  & \multicolumn{1}{c}{$\uparrow0.63\%$}
  & \multicolumn{1}{c}{$\uparrow4.39\%$} \\

    \midrule[0.8pt]
    \multicolumn{9}{c}{\textbf{Part 3.\; Estimation on Data Corrupted by Label Noise.}}\\
    \midrule[0.8pt]
    \multirow{4}{*}{Adult}
        & \multirow{2}{*}{$\rho = 0.1$}
& SGD-IE & $1.00(1.00)\times10^{-4}$ & 0.9758(0.00) & 0.9672(0.00) & 0.9721(0.00) & 0.9487(0.01) & 0.9206(0.03) \\
& & ACC-SGD-IE & \multicolumn{1}{l}{$\bm {1.00(1.00)\times10^{-4}}$} & \textbf{0.9809(0.00)} & \textbf{0.9742(0.01)} & \textbf{0.9786(0.00)} & \textbf{0.9514(0.02)} & \textbf{0.9361(0.03)} \\

        \cmidrule{2-9}
        & \multirow{2}{*}{$\rho = 0.3$}
& SGD-IE & $1.00(0.00)\times10^{-4}$ & 0.9882(0.00) & 0.9835(0.00) & \textbf{0.9884(0.00)} & 0.9675(0.02) & \textbf{0.9450(0.05)} \\
& & ACC-SGD-IE & \multicolumn{1}{l}{$\bm {1.00(0.00)\times10^{-4}}$} & \textbf{0.9884(0.00)} & \textbf{0.9846(0.00)} & 0.9818(0.00) & \textbf{0.9702(0.02)} & 0.9376(0.06) \\
    \midrule
    \multirow{4}{*}{20News}
        & \multirow{2}{*}{$\rho = 0.1$}
& SGD-IE & $5.80(12.60)\times10^{-3}$ & 0.9331(0.05) & 0.9423(0.04) & 0.8944(0.09) & 0.8493(0.13) & 0.7073(0.23) \\
& & ACC-SGD-IE & \multicolumn{1}{l}{$\bm {5.80(12.50)\times10^{-3}}$} & \textbf{0.9638(0.00)} & \textbf{0.9593(0.01)} & \textbf{0.9388(0.01)} & \textbf{0.9179(0.02)} & \textbf{0.8692(0.05)} \\
        \cmidrule{2-9}
        & \multirow{2}{*}{$\rho = 0.3$} 
& SGD-IE & $6.80(12.10)\times10^{-3}$ & 0.9032(0.06) & 0.9108(0.05) & 0.8493(0.10) & 0.7896(0.15) & 0.6450(0.21) \\
& & ACC-SGD-IE & \multicolumn{1}{l}{$\bm {6.30(12.20)\times10^{-3}}$} & \textbf{0.9637(0.00)} & \textbf{0.9570(0.01)} & \textbf{0.9421(0.02)} & \textbf{0.9134(0.04)} & \textbf{0.8557(0.06)} \\
    \midrule
    \multirow{4}{*}{MNIST}
        & \multirow{2}{*}{$\rho = 0.1$}
& SGD-IE & $5.00(7.00)\times10^{-4}$ & 0.9615(0.02) & 0.9504(0.02) & 0.9318(0.04) & 0.9096(0.06) & 0.8626(0.06) \\
& & ACC-SGD-IE & \multicolumn{1}{l}{$\bm {4.00(6.00)\times10^{-4}}$} & \textbf{0.9677(0.01)} & \textbf{0.9594(0.02)} & \textbf{0.9468(0.02)} & \textbf{0.9466(0.03)} & \textbf{0.9069(0.06)} \\
        \cmidrule{2-9}
        & \multirow{2}{*}{$\rho = 0.3$}
& SGD-IE & $4.00(7.00)\times10^{-4}$ &  0.9685(0.01) & 0.9663(0.02) & 0.9532(0.02) & 0.9178(0.02) & 0.8564(0.07) \\
& & ACC-SGD-IE & \multicolumn{1}{l}{$\bm {2.00(4.00)\times10^{-4}}$} & \textbf{0.9754(0.01)} & \textbf{0.9699(0.02)} & \textbf{0.9595(0.02)} & \textbf{0.9542(0.02)} & \textbf{0.8914(0.05)} \\
    \midrule
    \multicolumn{3}{c}{\textbf{Average Improve (\%)}}
    & \multicolumn{1}{c}{$\downarrow15.47\%$}
    & \multicolumn{1}{c}{$\uparrow1.98\%$}
    & \multicolumn{1}{c}{$\uparrow1.51\%$}
    & \multicolumn{1}{c}{$\uparrow3.03\%$}
    & \multicolumn{1}{c}{$\uparrow5.39\%$}
    & \multicolumn{1}{c}{$\uparrow10.94\%$} \\

    \specialrule{1pt}{0pt}{0pt}  
  \end{tabular}} 
\end{table*}

\subsection{Cross-Epoch Results across Batch Sizes and Learning Rates} \label{app:extended_cross_epoch}
We extend \autoref{subsec:cross-epoch-fidelity} to the complete set of tested
batch sizes and learning rates. In \autoref{fig:epoch-wise_both},
ACC-SGD-IE maintains lower RMSE for both convex and non-convex models.
Its Jaccard@10 is generally higher or comparable, showing that the
estimation benefit persists across these longer training trajectories.

\subsubsection{Convex-model experiments}

For the highlighted logistic-regression configuration ($M=400$,
$\alpha=1.0$), both estimators' RMSE decreases over training, with
ACC-SGD-IE maintaining the lower error. Its relative RMSE reduction
grows from $1\%$ at epoch~2 to $86\%$ at epoch~38. SGD-IE's Jaccard@10\%
falls to $0.4$, while ACC-SGD-IE stays above $0.6$. These convex controls
complement the non-convex comparison in
\autoref{subsec:cross-epoch-fidelity}.

In \autoref{fig:epoch-wise_both} (II), the configurations with $M=400$ show faster RMSE decay as the learning rate increases from $0.1$ to $0.3$ and $1.0$. Under the curvature and step-size conditions of \autoref{thm:strong}, the factor $1-\alpha(1-1/M)\lambda$ describes contraction of the ACC residual. The observed decay is consistent with this contraction mechanism. At $\alpha=0.01$, RMSE increases mildly over the measured horizon. At epoch~2, the RMSE in panel~(a) is $3\times10^{-2}$, compared with $4\times10^{-3}$ in panel~(d), illustrating the effect of batch size on the scale of the estimated loss change.

\subsubsection{Non-convex-model experiments}
ACC-SGD-IE reduces empirical RMSE relative to SGD-IE in all non-convex configurations shown in \autoref{fig:epoch-wise_both} (I). At $\alpha=0.01$, the relative reduction at epoch~38 is $32\%$, $36\%$, and $46\%$ for batch sizes $M=20$, $200$, and $400$, respectively. With $M=400$, the corresponding reductions are $46\%$, $43\%$, $42\%$, and $43\%$ for learning rates $0.01$, $0.1$, $0.3$, and $1.0$. Thus the correction improves estimation across the tested batch sizes and learning rates. These results complement \autoref{thm:nonconvex}, which isolates the additional interaction residual in SGD-IE and characterizes the Taylor residual retained by ACC-SGD-IE.

\section{Accumulative Correction for Related Estimators}
\label{sec:extension}

We empirically evaluate accumulative correction in related estimators, comparing ranking agreement for DVEmb and Adam-IE with and without ACC on MNIST.

\subsection{Experimental setup}
Experiments use \textsc{MNIST}, seeds 0--4, batch size 100, and learning rate 0.1, with means over five independent runs. We report TracIn and DVEmb rankings at epochs 4, 16, and 32, and Adam-IE rankings after 4 epochs. The implementations follow the formulations specified in the original papers.

The Adam variants follow the Adam-based update and cosine-similarity
formulation from prior work \citep{LESS}.

\subsection{Ranking Results}
The tables report ranking comparisons at the tested epochs and cutoffs.
These comparisons do not establish error bounds for DVEmb or Adam-IE.

\begin{table}[!htbp]
  \centering
  \small
  \setlength{\tabcolsep}{3.5pt}

  \caption{Ranking agreement for TracIn, DVEmb, and ACC-DVEmb across the evaluated epochs.}
  \label{tab:acc_dvemb}
  \begin{adjustbox}{max width=\linewidth}
  \begin{tabular}{@{}l l c c c c c@{}}
  \toprule
  Epoch & Method      & Kendall's Tau & Top70 & Top50 & Top30 & Top10 \\
  \midrule
  4  & TracIn        & 0.8167 & 0.7949 & 0.6598 & 0.6783 & 0.5686 \\
     & DVEmb         & 0.7772 & 0.7445 & 0.6529 & 0.6216 & 0.4815 \\
     & \textbf{ACC-DVEmb} & \textbf{0.8463} & \textbf{0.8123} & \textbf{0.7241} & \textbf{0.7021} & \textbf{0.5686} \\
  \midrule
  16 & TracIn        & 0.6705 & 0.6867 & 0.5564 & 0.4815 & 0.4815 \\
     & DVEmb         & 0.7073 & 0.7021 & 0.6260 & 0.5094 & 0.4545 \\
     & \textbf{ACC-DVEmb} & \textbf{0.7225} & \textbf{0.7125} & \textbf{0.6327} & \textbf{0.5190} & \textbf{0.5094} \\
  \midrule
  32 & TracIn        & $-0.1475$ & 0.5013 & 0.2862 & 0.1483 & 0.1111 \\
     & DVEmb         & 0.6662 & 0.6667 & 0.5504 & 0.5287 & 0.5094 \\
     & \textbf{ACC-DVEmb} & \textbf{0.7011} & \textbf{0.7125} & \textbf{0.6129} & \textbf{0.5686} & \textbf{0.5332} \\
  \bottomrule
  \end{tabular}
  \end{adjustbox}

\end{table}

\begin{table}[!htbp]
  \centering

  \small
  \setlength{\tabcolsep}{5pt}
  \caption{Jaccard overlap for Adam-IE and ACC-Adam-IE after four epochs.}
  \label{tab:acc_adam}
  \begin{tabular}{@{}l c c c c@{}}
  \toprule
  Method & J@70\% & J@50\% & J@30\% & J@10\% \\
  \midrule
  Adam-IE         & 0.5427 & 0.4085 & 0.1840 & 0.0390 \\
  \textbf{ACC-Adam-IE} & \textbf{0.5699} & \textbf{0.4337} & \textbf{0.2060} & \textbf{0.0390} \\
  \bottomrule
  \end{tabular}
\end{table}

\begin{figure*}[!t]
\centering
\input{appendix_cross_epoch_figure}
\end{figure*}

\bibliographystyle{IEEEtranN}  
\bibliography{reference}

\end{document}